\documentclass{article} % For LaTeX2e
\usepackage{iclr2017_conference,times}
\usepackage{url}
\usepackage{latexsym}
\usepackage{amsfonts}
\usepackage{graphicx}
\usepackage{amsmath}
\usepackage{amssymb}
\usepackage{booktabs}
\usepackage{tabularx}
\usepackage{multirow}
\usepackage{bbm}
\usepackage{algorithm}
\usepackage{subcaption}
\usepackage[font=small,labelfont=bf]{caption}
\usepackage[noend]{algpseudocode}
\usepackage{wrapfig}
\usepackage{todonotes}
\usepackage{tikz}
\usetikzlibrary{matrix,arrows,backgrounds,calc,patterns,positioning,fit,shapes}
\usepackage{tikz-dependency}
\usepackage[titletoc,title]{appendix}
\DeclareMathOperator*{\argmax}{argmax}

\DeclareMathOperator{\softmax}{softmax}
\DeclareMathOperator{\logadd}{logadd}
\DeclareMathOperator{\sign}{sign}
\DeclareMathOperator{\signexp}{signexp}
\DeclareMathOperator{\sigmoid}{sigmoid}

\DeclareMathOperator{\head}{head}

\DeclareMathOperator{\relu}{ReLU}
\DeclareMathOperator{\lstm}{LSTM}

\DeclareMathOperator{\mlp}{MLP}
\newcommand{\oplusgets}{\gets_{\oplus}}
\newcommand{\pgrad}{\nabla_{p}^\mathcal{L}}
\newcommand{\alphagrad}{\nabla_{\alpha}^\mathcal{L}}
\newcommand{\betagrad}{\nabla_{\beta}^\mathcal{L}}
\newcommand{\thetagrad}{\log \nabla_{\theta}^\mathcal{L}}

\newcommand{\xvec}{\mathbf{x}}
\newcommand{\yvec}{\mathbf{y}}
\newcommand{\cvec}{\mathbf{c}}
\newcommand{\mvec}{\mathbf{m}}

\newcommand{\qvec}{\mathbf{q}}

\newcommand{\svec}{\mathbf{s}}

\newcommand{\mcL}{\mathcal{L}}

\newcommand{\Uvec}{\mathbf{U}}

\newcommand{\E}{\mathbb{E}}

\newcommand{\Vvec}{\mathbf{V}}
\newcommand{\Wvec}{\mathbf{W}}
\newcommand{\hvec}{\mathbf{h}}

\newcommand{\bvec}{\mathbf{b}}
\newcommand{\reals}{\mathbb{R}}
\newcommand{\ind}{\mathbbm{1}}

\newcommand\given{\,|\,}

\title{Structured Attention Networks}
\author{Yoon Kim\thanks{Equal contribution.} \hspace{5mm} Carl Denton$^*$ \hspace{5mm} Luong Hoang  \hspace{5mm} Alexander M. Rush \\
\texttt{\small \{yoonkim@seas,carldenton@college,lhoang@g,srush@seas\}.harvard.edu}\\
School of Engineering and Applied Sciences\\
Harvard University\\
Cambridge, MA 02138, USA \\
}

\iclrfinalcopy % Uncomment for camera-ready version

\begin{document}

\maketitle

\begin{abstract}
  Attention networks have proven to be an effective approach for
  embedding categorical inference within a deep neural network.
  However, for many tasks we may want to model richer structural
  dependencies without abandoning end-to-end training. In this work,
  we experiment with incorporating richer structural distributions,
  encoded using graphical models, within deep networks. We
  show that these structured attention networks are simple extensions
  of the basic attention procedure, and that they allow for extending
  attention  beyond the standard soft-selection approach, such
  as attending to partial segmentations or to subtrees. We experiment
  with two different classes of structured attention networks: a
  linear-chain conditional random field and a graph-based parsing
  model, and describe how these models can be practically implemented
  as neural network layers. Experiments show that this approach is
  effective for incorporating structural biases, and structured attention
networks outperform baseline attention models on a variety of synthetic 
and real tasks: tree transduction, neural machine translation, question answering, and natural language
inference. We further find that models trained in this way learn 
interesting unsupervised hidden representations that generalize simple attention.

\end{abstract}

\section{Introduction}
Attention networks are now a standard part of the deep learning
toolkit, contributing to impressive results in neural machine
translation \citep{Bahdanau2015,Luong2015}, image captioning
\citep{Xu2015}, speech recognition \citep{Chorowski2015,Chan2015},
question answering \citep{Hermann2015,Sukhbaatar2015}, and
algorithm-learning \citep{Graves2014,Vinyals2015c}, among many other
applications (see \cite{Cho2015} for a comprehensive review).
This approach alleviates the bottleneck of compressing a source
into a fixed-dimensional vector by equipping a model with
variable-length memory \citep{Weston2014,Graves2014,Graves2016},
thereby providing random access into the source as needed. Attention
is implemented as a hidden layer which computes a categorical
distribution (or hierarchy of categorical distributions) to make a
soft-selection over source elements.

Noting the empirical effectiveness of attention networks, we also
observe that the standard attention-based architecture does not
directly model any \textit{structural dependencies} that may exist
among the source elements, and instead relies completely on the hidden
layers of the network.  While one might argue that these structural
dependencies can be learned implicitly by a deep model with enough
data, in practice, it may be useful to provide a structural
bias. Modeling structural dependencies at the final, \textit{output} layer has
been shown to be important in many deep learning applications, most
notably in seminal work on graph transformers \citep{LeCun1998}, key work on NLP 
\citep{Collobert2011}, and in many other areas
\citep[\textit{inter alia}]{Peng2009,Do2010,Jaderberg2014b,Chen2015b,Durrett2015,Lample2016}.

In this work, we consider applications which may require structural
dependencies at the attention layer, and develop \textit{internal}
structured layers for modeling these directly. This approach
generalizes categorical soft-selection attention layers by
specifying possible structural
dependencies in a soft manner.  Key applications will be the
development of an attention function that segments the source input
into subsequences and one that takes into account the latent
recursive structure (i.e. parse tree) of a source sentence.

Our approach views the attention mechanism as a graphical model over a
set of latent variables. The standard attention network can be seen
as an expectation of an annotation function with respect to a single
latent variable whose categorical distribution is parameterized to be
a function of the source.  In the general case we can specify a
graphical model over multiple latent variables whose edges encode the
desired structure. Computing forward attention requires performing
inference to obtain the expectation of the annotation function, i.e.
the \textit{context vector}. This expectation is computed over an
exponentially-sized set of structures (through the machinery of
graphical models/structured prediction), hence the name \textit{structured
  attention} network. Notably each step of this process (including inference)
is differentiable, so the model can be trained end-to-end without
having to resort to deep policy gradient methods \citep{schulman2015gradient}.

The differentiability of inference algorithms over graphical models
has previously been noted by various researchers
\citep{Li2009,Domke2011,Stoyanov2011,Stoyanov2012,Gormley2015},
primarily outside the area of deep learning. For example,
\citet{Gormley2015} treat an entire graphical model as a
differentiable circuit and backpropagate risk through variational
inference (loopy belief propagation) for minimium risk training of
dependency parsers. Our contribution is to combine these ideas to
produce structured \textit{internal} attention layers within deep
networks, noting that these approaches allow us to use the resulting
marginals to create new features, as long as we do so a differentiable
way.

We focus on two classes  of structured attention:
linear-chain conditional random fields (CRFs) \citep{Lafferty2001} and
first-order graph-based dependency parsers \citep{Eisner1996}. The
initial work of \cite{Bahdanau2015} was particularly interesting in
the context of machine translation, as the model was able to
implicitly learn an \textit{alignment model as a hidden layer},
effectively embedding inference into a neural network. In similar
vein, under our framework the model has the capacity to learn a
\textit{segmenter as a hidden layer} or a \textit{parser as a hidden
  layer}, without ever having to see a segmented sentence or a parse
tree. Our experiments apply this approach to a difficult synthetic
reordering task, as well as to machine translation, question answering, and 
natural language inference.  We find that models trained with structured attention
outperform standard attention models. Analysis
of learned representations further reveal that interesting structures
 emerge as an internal layer of the model. All code is available at 
\url{http://github.com/harvardnlp/struct-attn}.

\section{Background: Attention Networks}

A standard neural network consist of a series of non-linear
transformation layers, where each layer produces a fixed-dimensional
hidden representation. For tasks with large input spaces, this
paradigm makes it hard to control the interaction between
components. For example in machine translation, the source consists of
an entire sentence, and the output is a prediction for each word in
the translated sentence. Utilizing a standard network leads to an
information bottleneck, where one hidden layer must encode the entire
source sentence. Attention provides an alternative approach.\footnote{Another line
of work involves marginalizing over latent variables (e.g.
latent alignments) for sequence-to-sequence transduction 
\citep{Kong2016,Lu2016,Yu2016,Yu2017}.} An attention network
maintains a set of hidden representations that scale
 with the size
of the source. The model uses an internal
inference step to perform a soft-selection over these representations. 
This method allows the model to maintain a variable-length memory
and has shown to be crucially important for scaling systems for many tasks.

Formally, let $x = [x_1, \dots, x_n]$ represent a sequence of inputs,
let $q$ be a query, and let $z$ be a categorical latent variable with
sample space $\{1, \ldots, n\}$ that encodes the desired selection
among these inputs. Our aim is to produce a \textit{context} $c$ based
on the sequence and the query. To do so, we assume access to an
\textit{attention distribution} $z \sim p(z \given x, q)$, where we
condition $p$ on the inputs $x$ and a query $q$. The \textit{context}
over a sequence is defined as expectation,
$c = \E_{z \sim p(z \given x, q)} [f(x, z)]$ where $f(x, z)$ is an
\textit{annotation function}.  Attention of this form can be applied
over any type of input, however, we will primarily be concerned with
``deep'' networks, where both the annotation function and attention
distribution are parameterized with neural networks, and the context
produced is a vector fed to a downstream network.

For example, consider the case of attention-based neural machine
translation \citep{Bahdanau2015}.  Here the sequence of inputs
$[\mathbf{x}_1, \ldots, \mathbf{x}_n]$ are the hidden states of a
recurrent neural network (RNN), running over the words in the source
sentence, $\mathbf{q}$ is the RNN hidden state of the target decoder (i.e. vector
representation of the query $q$),
and $z$ represents the source position to be attended to for
translation. The attention distribution $p$ is simply
$p(z = i \given x, q) = \softmax(\theta_i)$ where $\theta \in \reals^n$
is a parameterized potential typically based on a neural network,
e.g. $\theta_i = \mlp([\mathbf{x}_i; \qvec])$.  The annotation function is
defined to simply return the selected hidden state,
$f(\mathbf{x}, z) = \mathbf{x}_z$.  The context vector can then be computed
using a simple sum,
\begin{equation}\label{vanilla-attn}
\mathbf{c} = \E_{z \sim p(z \given x, q)} [f( x, z)] = \sum_{i=1}^n p(z = i \given x, q) \mathbf{x}_i
\end{equation}
Other tasks such as question answering use attention in a similar
manner, for instance by replacing source $[x_1, \ldots, x_n]$ with a
set of potential facts and $q$ with a representation of the question.

In summary we interpret the attention mechanism as taking the
expectation of an annotation function $f(x,z)$ with respect to a latent
variable $z \sim p$, where $p$ is parameterized to be function of $x$
and $q$.

\section{Structured Attention}

Attention networks simulate selection from a set
using a soft model. In this work we consider generalizing selection to
types of attention, such as selecting chunks, segmenting inputs, or
even attending to latent subtrees. One interpretation of this
attention is as using soft-selection that considers all possible
structures over the input, of which there may be exponentially many
possibilities. Of course, this expectation can no longer be computed
using a simple sum, and we need to incorporate the machinery of inference
directly into our neural network.

Define a structured attention model as being an attention model where
$z$ is now a vector of discrete latent variables $[z_1, \ldots, z_m]$
and the attention distribution is $p(z \given x, q)$ is defined as a
\textit{conditional random field} (CRF), specifying the independence
structure of the $z$ variables. Formally, we assume an undirected
graph structure with $m$ vertices. The CRF is parameterized with
clique (log-)potentials $\theta_C(z_{C}) \in \reals$, where the $z_C$
indicates the subset of $z$ given by clique $C$.  Under this
definition, the attention probability is defined as,
$p(z \given x, q; \theta) = \softmax(\sum_C \theta_C(z_C))$, where for
symmetry we use $\softmax$ in a general sense,
i.e. $\softmax(g(z)) = \frac{1}{Z} \exp(g(z))$ where
$Z = \sum_{z'} \exp(g(z'))$ is the implied partition function. In
practice we use a neural CRF, where $\theta$ comes from a deep model
over $x, q$.

In structured attention, we also assume that the annotation function
$f$ factors (at least) into clique annotation functions
$f(x, z) = \sum_C f_C(x, z_C)$. Under standard conditions on the
conditional independence structure, inference techniques from
graphical models can be used to compute the forward-pass expectations
and the context:
\[c = \E_{z \sim p(z \given x, q)} [f(x, z)] = \sum_{C} \E_{z \sim p(z_C \given x, q)} [f_C(x, z_C)]\] 

\subsection{Example 1: Subsequence Selection}
\label{sec:subselect}

\begin{figure}
  \centering
  \begin{subfigure}{0.32\textwidth}
    \centering
  \begin{tikzpicture}
    \tikzstyle{latent} = [circle,fill=white,draw=black,inner sep=1pt,
    minimum size=20pt, font=\fontsize{10}{10}\selectfont]
    \tikzstyle{obs} = [latent,fill=gray!25]

    \node[obs](q){$q$};
    \node[below of=q, latent](za){$z_1$};
    \node[xshift= -1.5cm,below of=za, obs](xa){$x_1$};
    \node[right of=xa, obs](xb){$x_2$};
    \node[right of=xb, obs](xc){$x_3$};
    \node[right of=xc, obs](xd){$x_4$};

    \draw[dashed] (xa) to (za);
    \draw[dashed] (xb) to (za);
    \draw[dashed] (xc) to (za);
    \draw[dashed] (xd) to (za);
  \end{tikzpicture}
  \caption{}
  \end{subfigure}
  \begin{subfigure}{0.32\textwidth}
    \centering
  \begin{tikzpicture}
    \tikzstyle{latent} = [circle,fill=white,draw=black,inner sep=1pt,
    minimum size=20pt, font=\fontsize{10}{10}\selectfont]
    \tikzstyle{obs} = [latent,fill=gray!25]

    \node[obs](q){$q$};
    \node[xshift= -1.5cm, below of=q, latent](za){$z_1$};
    \node[right of=za, latent](zb){$z_2$};
    \node[right of=zb, latent](zc){$z_3$};
    \node[right of=zc, latent](zd){$z_4$};
    \node[below of=za, obs](xa){$x_1$};
    \node[below of=zb, obs](xb){$x_2$};
    \node[below of=zc, obs](xc){$x_3$};
    \node[below of=zd, obs](xd){$x_4$};

    \draw[dashed] (xa) to (za);
    \draw[dashed] (xb) to (zb);
    \draw[dashed] (xc) to (zc);
    \draw[dashed] (xd) to (zd);
  \end{tikzpicture}
  \caption{}
  \end{subfigure}
  \begin{subfigure}{0.3\textwidth}
    \centering
  \begin{tikzpicture}
    \tikzstyle{latent} = [circle,fill=white,draw=black,inner sep=1pt,
    minimum size=20pt, font=\fontsize{10}{10}\selectfont]
    \tikzstyle{obs} = [latent,fill=gray!25]

    \node[obs](q){$q$};
    \node[xshift= -1.5cm, below of=q, latent](za){$z_1$};
    \node[right of=za, latent](zb){$z_2$};
    \node[right of=zb, latent](zc){$z_3$};
    \node[right of=zc, latent](zd){$z_4$};
    \node[below of=za, obs](xa){$x_1$};
    \node[below of=zb, obs](xb){$x_2$};
    \node[below of=zc, obs](xc){$x_3$};
    \node[below of=zd, obs](xd){$x_4$};

    \draw[line width=1] (za) to (zb);
    \draw[line width=1] (zb) to (zc);
    \draw[line width=1] (zc) to (zd);

    \draw[dashed] (xa) to (za);
    \draw[dashed] (xb) to (zb);
    \draw[dashed] (xc) to (zc);
    \draw[dashed] (xd) to (zd);
  \end{tikzpicture}
  \caption{}
  \end{subfigure}
% \vspace{-50mm}
% \includegraphics[scale=1]{tikz1.pdf}
% \vspace{-50mm}
  \caption{\small \label{fig:seq} Three versions of a latent variable attention
    model:
 (a) A standard soft-selection attention network, (b) A
    Bernoulli (sigmoid) attention network, (c) A linear-chain
    structured attention model for segmentation. The input and query are denoted with $x$ and $q$ respectively. }
\end{figure}
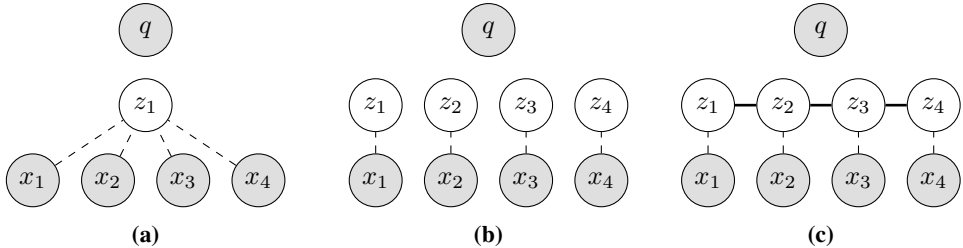

Suppose instead of soft-selecting a single input, we wanted to
explicitly model the selection of contiguous subsequences. We could
naively apply categorical attention over all subsequences, or hope
the model learns a multi-modal distribution to combine neighboring
words. Structured attention provides an alternate approach.

Concretely, let $m =n$, define $z$ to be a random vector
$z = [z_1, \dots, z_n]$ with $z_i \in \{0, 1\}$, and define our
annotation function to be, $f(x,z) = \sum_{i=1}^n f_{i} (x,z_{i})$ where $f_{i} (x,z_i) =  \ind \{ z_i = 1\} \xvec_i$.
The explicit expectation is then,
\begin{equation}\label{struct-attn}
\E_{z_1, \dots, z_n }[f(x,z)] = \sum_{i=1}^n p(z_i = 1 \given x, q) \xvec_i
\end{equation}

Equation (\ref{struct-attn}) is similar to equation (\ref{vanilla-attn})---both are a linear
combination of the input representations where the scalar is between $[0,1]$ and represents how much
attention should be focused on each input. However, (2) is fundamentally different in two ways: (i) 
it allows for multiple inputs (or no inputs) to be selected for a given query;  (ii)
we can incorporate structural dependencies across the $z_i$'s. For instance, we can model the distribution over $z$ with a linear-chain CRF with pairwise edges,
\begin{align}\label{linear-chain}
p(z_1, \dots, z_n \given x, q) = \softmax \left( \sum_{i=1}^{n-1}   \theta_{i,i+1}(z_i, z_{i+1})  \right)
\end{align}
where $\theta_{k,l}$ is the pairwise potential for $z_i = k$ and $z_{i+1} = l$. This model is shown in Figure~\ref{fig:seq}c. 
Compare this model to the standard attention in Figure~\ref{fig:seq}a, or to a simple Bernoulli (sigmoid) selection 
method, $p(z_i = 1 \given x, q) = \sigmoid(\theta_{i}) $, shown in Figure~\ref{fig:seq}b.
 All three of these methods can use potentials 
from the same neural network or RNN that takes $x$ and $q$ as inputs.

In the case of the linear-chain CRF in~(\ref{linear-chain}), the marginal distribution
$p(z_i = 1 \given x)$ can be calculated efficiently in
linear-time for all $i$ using message-passing, i.e. the
forward-backward algorithm.  These marginals allow us to calculate
(\ref{struct-attn}), and in doing so we implicitly sum over an
exponentially-sized set of structures (i.e. all binary sequences of
length $n$) through dynamic programming. 
We refer to this type of attention layer as a \emph{segmentation attention} layer.

Note that the forward-backward algorithm is being used as parameterized
\textit{pooling} (as opposed to output computation), and can be
thought of as generalizing the standard attention softmax.
Crucially this generalization from vector softmax to forward-backward
is just a series of differentiable steps,\footnote{As are other dynamic
  programming algorithms for inference in graphical models, such as
  (loopy and non-loopy) belief propagation.} and we can compute
gradients of its output (marginals) with respect to its input
(potentials). This will allow the structured attention model to be
trained end-to-end as part of a deep model.

\subsection{Example 2: Syntactic Tree Selection }
This same approach can be used for more involved structural
dependencies.  One popular structure for natural language tasks is a
dependency tree, which enforces a structural bias on the recursive
dependencies common in many languages. In particular a dependency tree
enforces that each word in a source sentence is assigned exactly one
parent word (\textit{head word}), and that these assignments do not cross (projective
structure). Employing this bias encourages the system to make a
soft-selection based on learned syntactic dependencies, without
requiring linguistic annotations or a pipelined decision.

A dependency parser can be partially formalized as a graphical model with the
following cliques \citep{Smith2008}: latent variables
$z_{ij} \in \{0,1\}$ for all $i \ne j$, which indicates that the
$i$-th word is the parent of the $j$-th word (i.e.
$x_i \rightarrow x_j$); and a special global constraint that rules out
configurations of $z_{ij}$'s that violate parsing constraints
(e.g. one head, projectivity). 

The parameters to the graph-based CRF dependency parser are the
potentials $\theta_{ij}$, which reflect the score of selecting
$x_i$ as the parent of $x_j$. The probability of a parse tree $z$
given the sentence $x = [x_1, \ldots, x_n]$ is,
\begin{equation}
p(z \given x, q)= \softmax \left(\ind\{z\  \text{is valid}\}  \sum_{i \neq j}  \ind\{z_{ij} = 1\} \theta_{ij} \right)
\end{equation}
where $z$ is represented as a vector of $z_{ij}$'s for all $i \ne j$. 
It is possible to calculate the marginal probability of each edge $p(z_{ij} = 1\given x, q)$
for all $i, j$ in $O(n^3)$ time using the inside-outside algorithm \citep{Baker1979}
on the data structures of \citet{Eisner1996}. 

The parsing contraints ensure that each word has exactly one head (i.e. $\sum_{i=1}^n z_{ij} = 1$). Therefore if we want to utilize the \emph{soft-head}
 selection of a position $j$, 
the context vector is defined as: 
\begin{align*}
f_j(x, z) = \sum_{i=1}^n \ind\{z_{ij} = 1\} \xvec_i &  & \cvec_j = \E_z [f_j(x, z)] = \sum_{i=1}^n p(z_{ij} = 1 \given x, q) \xvec_i
\end{align*}
Note that in this case the annotation function has the subscript $j$ to produce a context vector for each word in the sentence.
Similar types of attention can be applied for other tree properties (e.g. soft-children).
We refer to this type of attention layer as a \emph{syntactic attention} layer.

\subsection{End-to-End Training}\label{sec:e2e}

\begin{figure}
\small
  \begin{subfigure}[t]{0.45\textwidth}
    \begin{algorithmic}  
      \Procedure{ForwardBackward}{$\theta$}
      \State{$\alpha[0, \langle t \rangle] \gets 0$} 
      \State{$\beta[n+1, \langle t\rangle] \gets 0$} 
      \For{$i =  1, \dots , n; c \in \mathcal{C}$}
      \State{$\alpha[i, c] \gets \bigoplus_{y} \alpha[i-1, y] \otimes \theta_{i-1, i}(y, c) $}
      \EndFor
      \For{$i  = n, \dots, 1; c \in \mathcal{C}$}
      \State{$\beta[i, c] \gets \bigoplus_{y} \beta[i+1, y] \otimes \theta_{i,i+1}(c, y) $}
      \EndFor
      \State{$A \gets \alpha[n+1, \langle t\rangle]$}
      \For{$i = 1, \dots, n; c \in \mathcal{C}$}
      \State{$ p(z_i = c \given x) \gets \exp (\alpha[i, c] \otimes \beta[i, c]$}
      \State{\hspace{18mm}$\otimes -A)$}
      \EndFor
      \State{\Return{$p$}}
      \EndProcedure{}
    \end{algorithmic}
  \end{subfigure}
  \begin{subfigure}[t]{0.6\textwidth}
    \begin{algorithmic}  
      \Procedure{BackpropForwardBackward}{$\theta, p, \nabla^\mathcal{L}_{p}$}  
      \State{$\nabla^\mathcal{L}_\alpha \gets \log p \otimes \log \pgrad \otimes \beta \otimes - A$}
      \State{$\nabla^\mathcal{L}_\beta \gets  \log p \otimes \log \pgrad \otimes \alpha \otimes - A$} 

      \State{$\hat \alpha[0, \langle t \rangle] \gets 0$}
      \State{$\hat \beta[n+1, \langle t \rangle] \gets 0$} 
      \For{$i = n, \dots 1;  c \in \mathcal{C}$}
      \State{$\hat\beta[i, c] \gets \nabla^\mathcal{L}_{\alpha}[i,c] \oplus \bigoplus_y \theta_{i,i+1}(c, y) \otimes \hat \beta[i+1, y]$}
      \EndFor

      \For{$i = 1, \dots, n; c \in \mathcal{C}$}
      \State{$\hat \alpha[i, c] \gets \nabla^\mathcal{L}_{\beta}[i,c] \oplus \bigoplus_y \theta_{i-1,i}(y, c) \otimes \hat \alpha[i-1, y]$}
      \EndFor

      \For{$i = 1, \dots, n; y, c \in \mathcal{C}$}
      \State{$\nabla^\mathcal{L}_{\theta_{i-1,i}(y, c)} \gets \signexp (\hat \alpha[i, y] \otimes \beta[i+1, c] $} 
      \State{\hspace{2.8pc} $\oplus\: \alpha[i, y] \otimes \hat \beta[i+1, c]$}
      \vspace{.3pc}
      \State{\hspace{2.8pc} $\oplus\: \alpha[i, y] \otimes \beta[i+1, c] \otimes -A)$} 
      \EndFor
      \State{\Return{$\nabla^\mathcal{L}_{\theta}$}}
      
      \EndProcedure{}
    \end{algorithmic}
  \end{subfigure}
  \caption{\label{fig:fb} \small Algorithms for linear-chain CRF: (left)
computation of forward-backward tables $\alpha$, $\beta$, and marginal probabilities $p$
    from potentials $\theta$ (forward-backward algorithm); (right) 
backpropagation of loss gradients with respect to the marginals $\nabla_{p}^{\cal L}$.
 $\mathcal{C}$ denotes the state space and $\langle t\rangle$ is the special start/stop state. Backpropagation uses the identity $\nabla_{\log p}^\mcL = p \odot \pgrad$ to 
calculate $\nabla^\mcL_{\theta} = \nabla_{\log p}^\mcL \nabla^{\log p}_{\theta}$, where $\odot$ is the element-wise multiplication.
Typically the forward-backward with marginals is performed in the
    log-space semifield $\mathbb{R}\cup\{\pm \infty\}$ with binary operations $\oplus = \logadd$ and  $\otimes = +$ for numerical precision. However, 
    backpropagation requires working with the log of negative values (since $\pgrad$ could be negative), 
so we extend to a field $\left[\mathbb{R}\cup\{\pm \infty\}\right] \times \{+, -\}$ with special $+$/$-$ 
log-space operations. Binary operations applied to vectors are implied to be element-wise. The
$\signexp$ function is defined as $\signexp(l_a) = s_a \exp (l_a)$.
See Section~\ref{sec:e2e} and Table~\ref{tab:dlog} for more details.
  }
\end{figure}

Graphical models of this form have been widely used as the final layer
of deep models. Our contribution is to argue that these networks can
be added within deep networks in place of simple attention layers. The
whole model can then be trained end-to-end.

The main complication in utilizing this approach within the network
itself is the need to backpropagate the gradients through an inference algorithm as
part of the structured attention network. Past work has demonstrated 
the techniques necessary for this approach (see \citet{Stoyanov2011}), but to our knowledge it
is very rarely employed.

Consider the case of the simple linear-chain CRF layer from equation (\ref{linear-chain}). 
Figure~\ref{fig:fb} (left) shows the standard
forward-backward algorithm for computing the marginals $p(z_i = 1\given x, q; \theta)$.
If we treat the forward-backward algorithm as a neural network layer, its input are the
potentials $\theta$, and its output after the forward pass are
these marginals.\footnote{Confusingly, ``forward'' in this case is
  different than in the \textit{forward}-backward algorithm, as the marginals
  themselves are the output. However the two uses of the term are
  actually quite related. The forward-backward algorithm can be
  interpreted as a forward and backpropagation pass on the log partition
  function. See \citet{Eisner2016} for further details
  (appropriately titled ``Inside-Outside and Forward-Backward
  Algorithms Are Just Backprop''). As such our full approach can be
  seen as computing second-order information. This interpretation is
  central to \citet{Li2009}.}  To backpropagate a loss through this
layer we need to compute the gradient of the loss $\mcL$ with respect to
$\theta$, $\nabla_{\theta}^\mcL$, as a function of the gradient
of the loss with respect to the marginals, $\nabla_{p}^\mcL$.\footnote{In general
we use $\nabla^a_b$ to denote the Jacobian of $a$ with respect to $b$.} As the
forward-backward algorithm consists of differentiable steps, this
function can be derived using reverse-mode automatic differentiation
of the forward-backward algorithm itself. Note that this reverse-mode algorithm
conveniently has a parallel structure to the forward version, and can
also be implemented using dynamic programming. 

\begin{wraptable}{r}{0.54\textwidth}
  \small
  \centering
  \begin{tabular}{cc|cc|cc}
    \toprule
   & & \multicolumn{2}{c|}{$\oplus$} & \multicolumn{2}{c}{$\otimes$} \\ 
    $s_a$ & $s_b$ &  $ l_{a+b} $  & $s_{a+b}$ & $ l_{a\cdot b}$ & $s_{a \cdot b}$\\
    \midrule
    $+$ & $+$ & $l_a+\log (1 + d)$&  $+$ & $l_a+l_b$ &$+$ \\ 
    $+$ & $-$ & $l_a+\log (1 - d)$& $+$ & $l_a+l_b$ &$-$ \\
    $-$ & $+$ & $l_a+\log (1 - d)$& $-$ & $l_a+l_b$ &$-$ \\
    $-$ & $-$ & $l_a+\log (1 + d)$& $-$ & $l_a+l_b$ &$+$ \\ 
    \bottomrule
  \end{tabular}
  \caption{\label{tab:dlog} \small Signed log-space semifield (from \cite{Li2009}). Each real number $a$
 is represented as a pair $( l_a, s_a )$ where $l_a = \log |a|$ and $s_a = \sign(a)$.
Therefore $a = s_a \exp(l_a)$. For the above we let $d = \exp(l_b- l_a)$ and assume $|a| > |b|$.
}
\end{wraptable}

However, in practice, one cannot simply use current off-the-shelf
tools for this task. For one, efficiency is quite important for these
models and so the benefits of hand-optimizing the reverse-mode
implementation still outweighs simplicity of automatic
differentiation. Secondly, numerical precision becomes a major issue
for structured attention networks. For computing the forward-pass and
the marginals, it is important to use the standard log-space semifield
over  $\mathbb{R}\cup\{\pm \infty\}$ with binary operations 
$(\oplus = \logadd, \otimes = +)$ to avoid underflow of probabilities. For computing
the backward-pass, we need to remain in log-space, but also handle 
log of negative values (since $\pgrad$ could be negative). This requires extending to the 
\textit{signed} log-space semifield over $\left[\mathbb{R}\cup\{\pm \infty\}\right] \times \{+, -\}$
with special $+$/$-$ operations. Table~\ref{tab:dlog}, based on \cite{Li2009},
demonstrates how to handle this issue, and Figure~\ref{fig:fb} (right)
describes backpropagation through the forward-backward algorithm.
For dependency parsing,
the forward pass can be computed using the
inside-outside implementation of Eisner's algorithm
\citep{Eisner1996}. Similarly, the backpropagation parallels the inside-outside
structure. Forward/backward pass through the inside-outside algorithm is
described in Appendix~\ref{app:io}.

\section{Experiments}

We experiment with three instantiations of structured attention networks on
four different tasks: (a) a simple, synthetic tree manipulation task using
the syntactic attention layer,
(b) machine translation with segmentation attention (i.e. two-state linear-chain CRF), 
(c) question answering using an $n$-state linear-chain CRF for 
multi-step inference over $n$ facts, and (d) 
natural language inference with syntactic tree attention.
These experiments are not intended to boost the state-of-the-art for
these tasks but to test whether these methods can be trained
effectively in an end-to-end fashion, can yield improvements over
standard selection-based attention, and can learn plausible latent
structures. All model architectures, hyperparameters, and training details
are further described in Appendix~\ref{app:model}.

\subsection{Tree Transduction}

The first set of experiments look at a tree-transduction task. These
experiments use synthetic data to explore a failure case of
soft-selection attention models. The task is to learn to convert a
random formula given in prefix notation to one in infix notation,
e.g.,
\begin{small}
\begin{align*}
(\,\,\,*\,\,\,(\,\,\,+\,\,\,(\,\,\,+\,\,\,15\,\,\,7\,\,\,)\,\,\,1\,\,\,8\,\,\,)\,\,\,(\,\,\,+\,\,\,19\,\,\,0\,\,\,11\,\,\,)\,\,\,) \,\, \Rightarrow
(\,\,(\,\,15\,\,+\,\,7\,\,\,)\,\,+\,\,1\,\,+\,\,8\,\,\,)\,\,*\,\,(\,\,\,19\,\,+\,\,0\,\,+\,\,11\,\,\,)
\end{align*}
\end{small}
The alphabet consists of symbols $\{(, ),+,*\}$, numbers between $0$
and $20$, and a special root symbol $\$$. This task is used
as a preliminary task to see if the model is able to learn the implicit
tree structure on the source side. The
model itself is an encoder-decoder model, where the encoder is defined below
and the decoder is an LSTM. See
Appendix~\ref{app:tree} for the full model.

Training uses $15$K prefix-infix pairs where the maximum nesting depth
is set to be between $2$-$4$ (the above example has depth $3$), with
$5$K pairs in each depth bucket. The number of expressions in each
parenthesis is limited to be at most $4$.  Test uses $1$K unseen
sequences with depth between $2$-$6$ (note specifically deeper than
train), with $200$ sequences for each depth. The performance is
measured as the average proportion of correct target tokens produced
until the first failure (as in \cite{Grefenstette2015}).

For experiments we try using different forms of \textit{self}-attention over embedding-only encoders.  
Let $\mathbf{x}_j$ be an embedding for each source symbol; our three
variants of the source representation $\hat{\xvec}_j$ are: (a) \textit{no atten}, just symbol
embeddings by themselves, i.e. $\hat{\xvec}_j = \mathbf{x}_j$; (b) \textit{simple} attention, symbol
embeddings and soft-pairing for each symbol, i.e. $ \hat{\xvec}_j =
[\mathbf{x}_j; \mathbf{c}_j]$ where $ \mathbf{c}_j = \sum_{i=1}^n
\softmax( \theta_{ij}) \mathbf{x}_i$ is calculated using soft-selection; (c)
\textit{structured} attention, symbol embeddings and soft-parent,
i.e. $\hat{\xvec}_j = [\mathbf{x}_j; \mathbf{c}_j]$ where $
\mathbf{c}_j = \sum_{i=1}^n p(z_{ij} = 1 \given x) \mathbf{x}_i $ is
calculated using parsing marginals, obtained from the syntactic attention layer. 
None of these models use an 
explicit query value---the potentials come from running a bidirectional 
LSTM over the source, producing hidden vectors $\hvec_i$, and then computing  \[\theta_{ij} = \tanh(\mathbf{s}^\top
\tanh(\mathbf{W}_1 \mathbf{h}_i + \mathbf{W}_2 \mathbf{h}_j + \mathbf{b}))\] 
\noindent where $\mathbf{s}, \mathbf{b}, \mathbf{W}_1, \mathbf{W}_2$ are parameters (see Appendix~\ref{app:parsing}).

\begin{wraptable}{l}{0.43\textwidth}\label{tree-perf}
\small
\begin{tabular}{c ccc} 
\toprule
Depth & No Atten  & Simple & Structured \\
\midrule
$2$ & $7.6$ & $87.4$ & $99.2$ \\
$3$ & $4.1$ & $49.6$ & $87.0$ \\
$4$ & $2.8$ & $23.3$ & $64.5$ \\
$5$ & $2.1$ & $15.0$ & $30.8$ \\
$6$ & $1.5$ & $8.5$ & $18.2$  \\
\bottomrule
\end{tabular}
\caption{\label{tree-perf} \small Performance (average length to failure \%) of models on the tree-transduction task.}
\end{wraptable}

\begin{figure}\label{tree-viz}
\centering
\begin{subfigure}{.49\textwidth}
  \centering
  \includegraphics[width=0.8\linewidth]{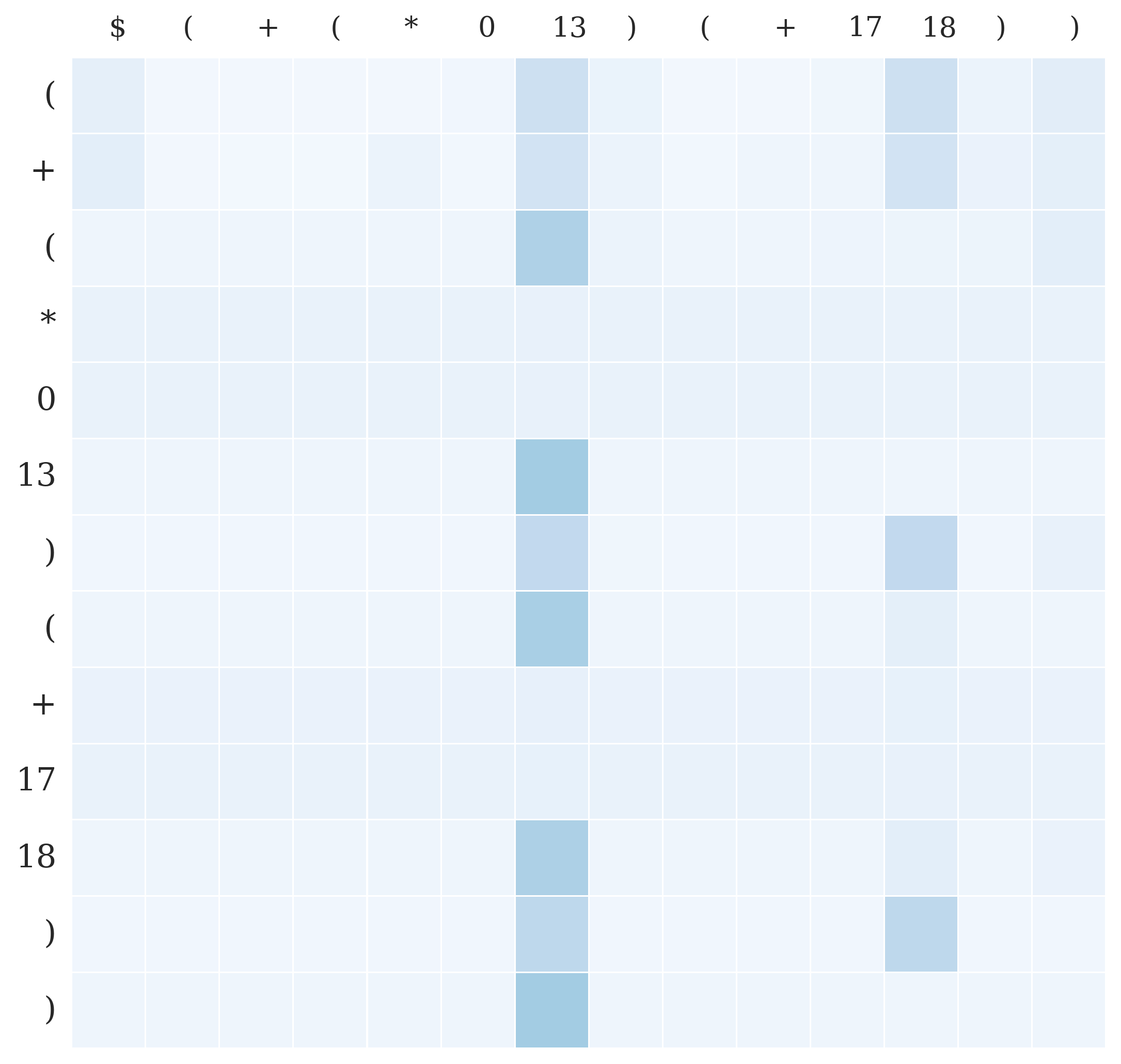}
\end{subfigure}
\begin{subfigure}{.49\textwidth}
  \centering
  \includegraphics[width=0.8\linewidth]{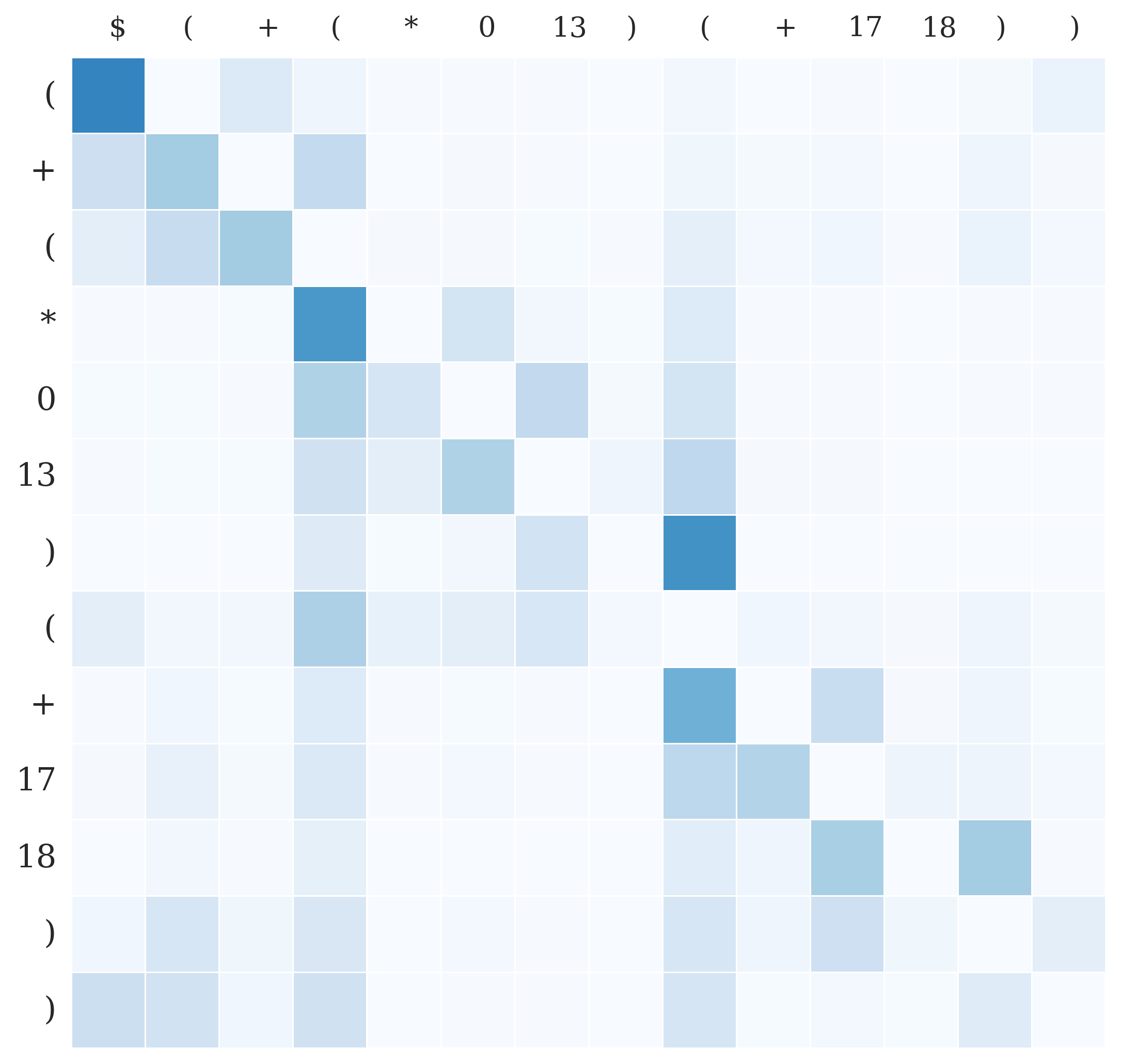}
\end{subfigure}
\caption{\label{tree-viz} \small Visualization of the source self-attention distribution for the simple (left) and structured (right) attention
models on the tree transduction task. $\$$ is the special root symbol. Each row delineates the distribution
over the parents (i.e. each row sums to one). The attention distribution obtained from the parsing marginals 
are more able to capture the tree structure---e.g. the attention weights of
closing parentheses are generally placed on the opening parentheses (though not necessarily on a single parenthesis).
 }
\end{figure}

The source representation $[\hat{\xvec}_1, \dots, \hat{\xvec}_n]$ are attended over using the standard attention mechanism at each
decoding step by an LSTM decoder.\footnote{Thus there are two attention mechanisms at work
  under this setup. First, structured attention over the source only to obtain
  soft-parents for each symbol (i.e. self-attention). Second, standard softmax alignment attention
  over the source representations during decoding.} 
Additionally, symbol
embedding parameters are shared between the parsing LSTM and the source
encoder.

\paragraph{Results}
Table~\ref{tree-perf} has the results for the task. Note that this task is 
fairly difficult as the encoder is quite simple.
The baseline model (unsurprisingly) performs poorly
as it has no information about the source ordering. The simple attention model  performs better, but is significantly outperformed by the structured model with a tree structure bias. We hypothesize that the model is partially reconstructing the arithmetic tree.  Figure~\ref{tree-viz} shows the attention distribution for the simple/structured models on 
the same source sequence, which indicates that the structured model is able to learn boundaries (i.e. parentheses).

\subsection{Neural Machine Translation}
Our second set of experiments use a full neural machine translation
model utilizing attention over subsequences. Here both the encoder/decoder are LSTMs, and 
we replace standard simple attention with a segmentation
attention layer. We experiment with two settings: translating directly from
unsegmented Japanese characters to English words (effectively using structured attention to perform soft word segmentation), 
and translating from segmented Japanese words to English words (which can be interpreted as doing \emph{phrase-based} 
neural machine translation). Japanese word segmentation is done using
the KyTea toolkit \citep{Neubig2011}. 

The data comes from the Workshop on Asian Translation (WAT) \citep{wat2016}.
We randomly pick $500$K sentences from the original training set (of $3$M sentences)
where the Japanese sentence was at most $50$ characters and the English sentence was at most
 $50$ words. We apply the same length filter on the provided validation/test sets for evaluation.
The vocabulary consists of all tokens that occurred
at least $10$ times in the training corpus. 

The segmentation attention layer is a two-state CRF where the unary potentials
at the $j$-th decoder step are parameterized as
\[
    \theta_i(k)= 
\begin{cases}
    \hvec_i \Wvec \hvec_j,& k = 1           \\ 0,  &k = 0
\end{cases}
\]
Here $[\hvec_1, \dots, \hvec_n]$ are the encoder hidden states and 
$\mathbf{h}_j'$ is the $j$-th decoder hidden state (i.e. the query vector).
 The pairwise potentials are
parameterized linearly with $\mathbf{b}$, i.e.  all together
\[ \theta_{i,i+1}(z_i, z_{i+1}) = \theta_i(z_i) + \theta_{i+1}(z_{i+1})
+ \mathbf{b}_{z_i, z_{i+1}} \] Therefore the segmentation attention layer requires
just $4$ additional parameters. Appendix~\ref{app:nmt} describes the full model
architecture.

We experiment with three attention configurations: (a) 
standard {\it simple} attention, i.e.
$\cvec_j = \sum_{i=1}^n \softmax(\theta_i) \hvec_i $; (b)
\textit{sigmoid} attention: multiple selection with Bernoulli random variables,
i.e. $\cvec_j = \sum_{i=1}^n \sigmoid(\theta_i) \hvec_i$; (c)
\textit{structured} attention, encoded with normalized CRF marginals,
\begin{align*}
\cvec_j = \sum_{i=1}^n \frac{p(z_i=1 \given x, q)}{\gamma} \hvec_i  & & \gamma = \frac{1}{\lambda} \sum_{i=1}^n p(z_i =1 \given x, q)  
\end{align*}
The normalization term $\gamma$ is not ideal but we found it to be helpful for stable 
training.\footnote{With standard expectation (i.e. $\cvec_j = \sum_{i=1}^n p(z_i=1 \given x, q) \hvec_i$)
we empirically observed the marginals to quickly saturate. 
We tried various strategies to overcome this, such as putting an $l_2$ penalty on the unary potentials and initializing
with a pretrained sigmoid attention model, but simply normalizing the marginals proved to be the most effective.
However, this changes the interpretation of the context vector as the expectation of an annotation function in this case.} 
$\lambda$ is a hyperparameter (we use $\lambda = 2$) and we 
further add an $l_2$ penalty of $0.005$ on the pairwise potentials $\bvec$. These values 
were found via grid search on the validation set.

\begin{wraptable}{l}{0.43\textwidth}\label{nmt-perf}
\small
\begin{tabular}{c ccc} 
\toprule
& Simple & Sigmoid & Structured \\
\midrule
\textsc{Char} & $12.6$ & $13.1$ & $14.6$ \\
\textsc{Word} & $14.1$ & $13.8$ & $14.3$ \\
\bottomrule
\end{tabular}
\caption{\label{nmt-perf}\small Translation performance as measured by BLEU (higher is better)
 on character-to-word and word-to-word Japanese-English translation for the three different models.}
\end{wraptable}

\paragraph{Results}
Results for the translation task on the test set are given in
Table~\ref{nmt-perf}.
Sigmoid attention outperforms simple (softmax) attention
 on the character-to-word task, potentially because it is able to learn many-to-one alignments.
On the word-to-word task, the opposite is true, with simple attention outperforming sigmoid attention.
Structured attention outperforms both models on both tasks, although improvements on the word-to-word
task are modest and unlikely to be statistically significant.

\begin{figure}[h]\label{nmt-viz}
\begin{subfigure}[t]{.5\textwidth}
\centering
\includegraphics[width=\linewidth]{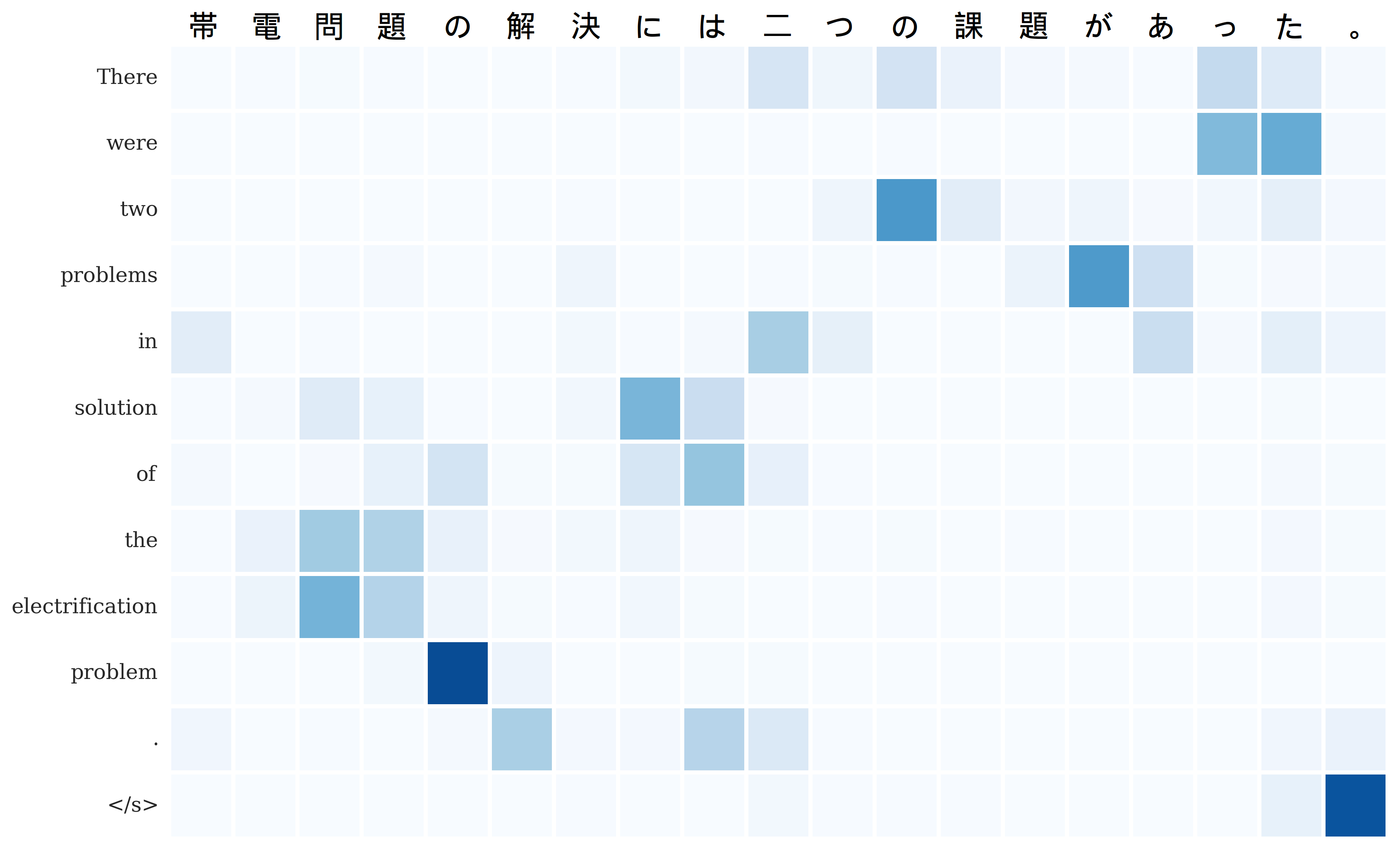}
\end{subfigure}
\begin{subfigure}[t]{.5\textwidth}
\centering
\includegraphics[width=\linewidth]{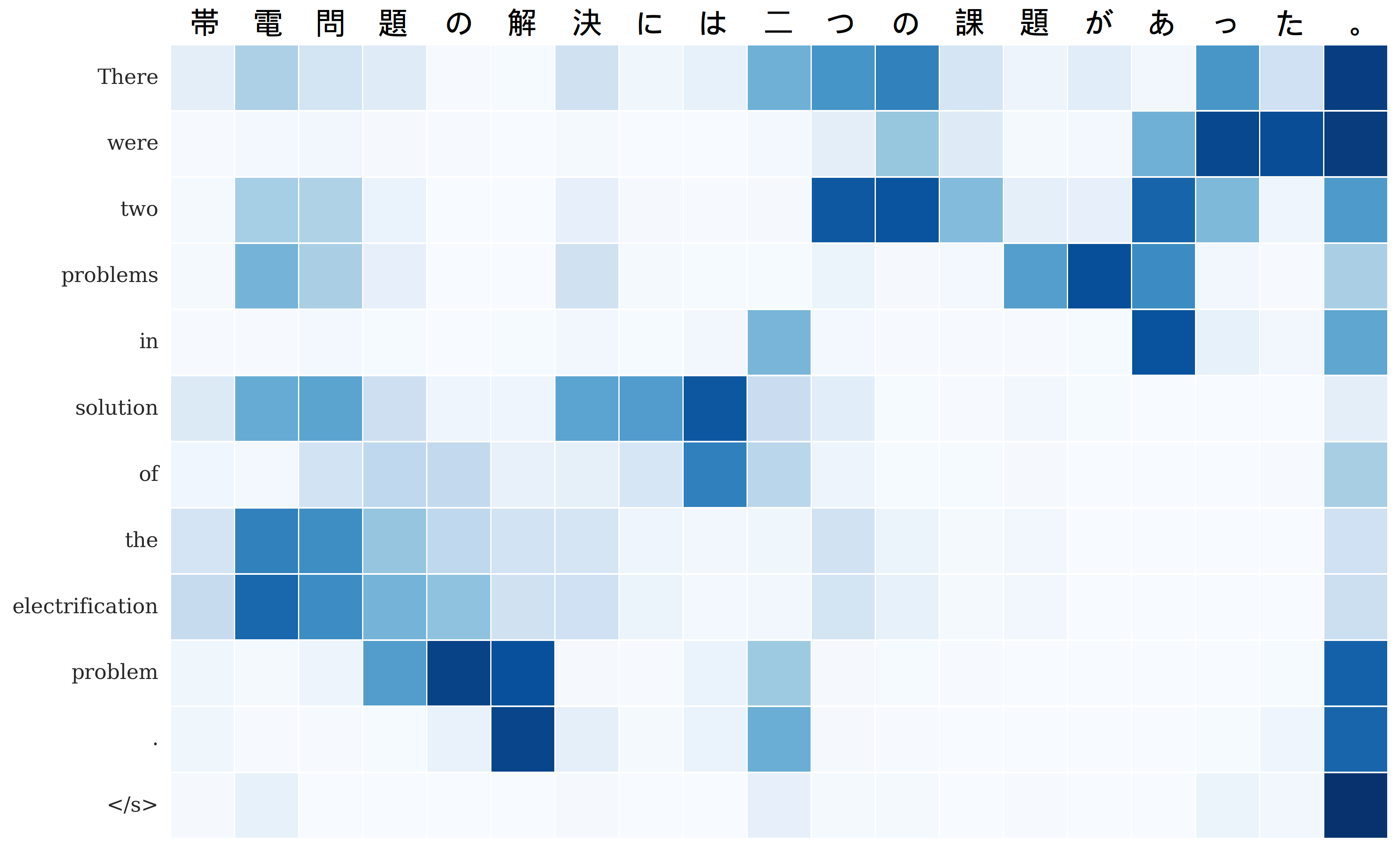}
\end{subfigure}
\medskip
\begin{subfigure}[t]{.5\textwidth}
\centering
\includegraphics[width=\linewidth]{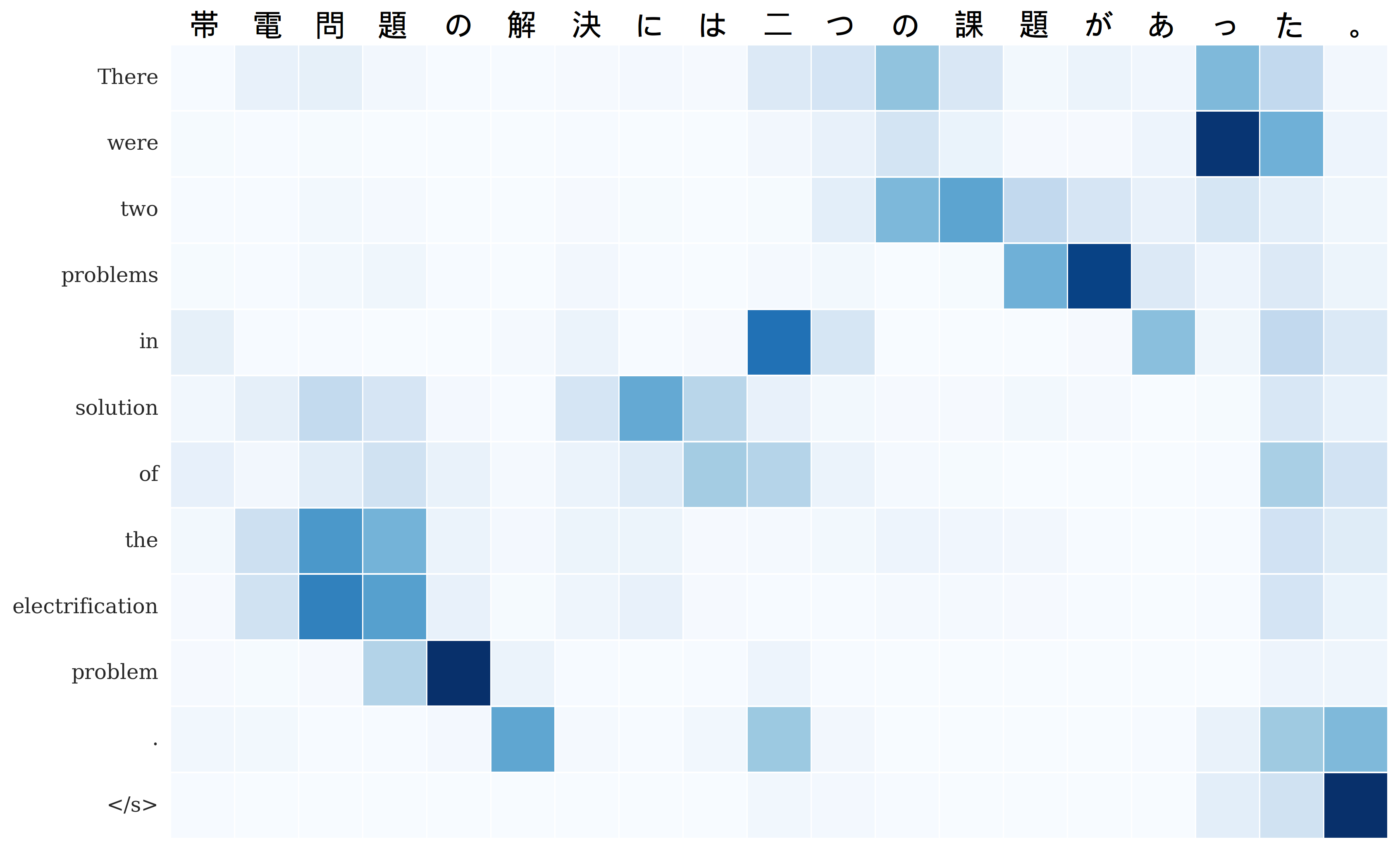}
\end{subfigure}
\medskip 
\begin{subfigure}[t]{.5\textwidth}
\centering
\includegraphics[width=\linewidth]{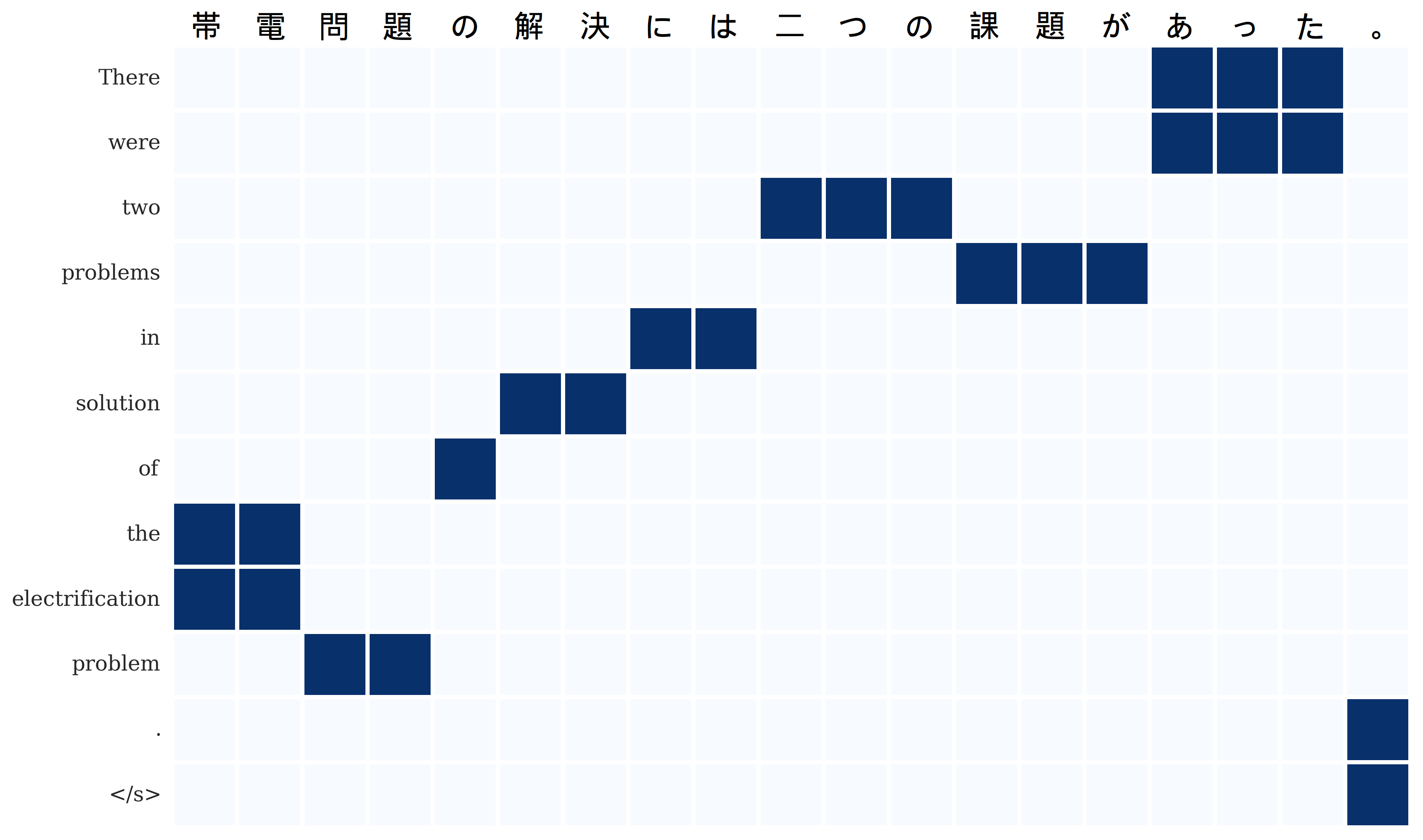}
\end{subfigure}

\caption{\label{fig:vis3} \small Visualization of the source attention distribution for the simple (top left), sigmoid (top right), and structured (bottom left) attention
models over the ground truth sentence on the character-to-word translation task. Manually-annotated alignments are shown in
bottom right. Each row delineates the attention weights over the source sentence at each step of
decoding. The sigmoid/structured attention models are able learn an implicit segmentation model
and focus on multiple characters at each time step.}
\end{figure}

For further analysis, Figure~\ref{fig:vis3} shows a visualization of
the different attention mechanisms on the character-to-word setup.  The simple model generally focuses attention heavily
on a single character. In contrast, the sigmoid and structured models are able to
spread their attention distribution on contiguous subsequences.
 The structured attention learns additional parameters (i.e. $\bvec$) to smooth out this type of attention.
\subsection{Question Answering}
\begin{figure}\label{qa-heat}
\centering
\includegraphics[scale=0.24]{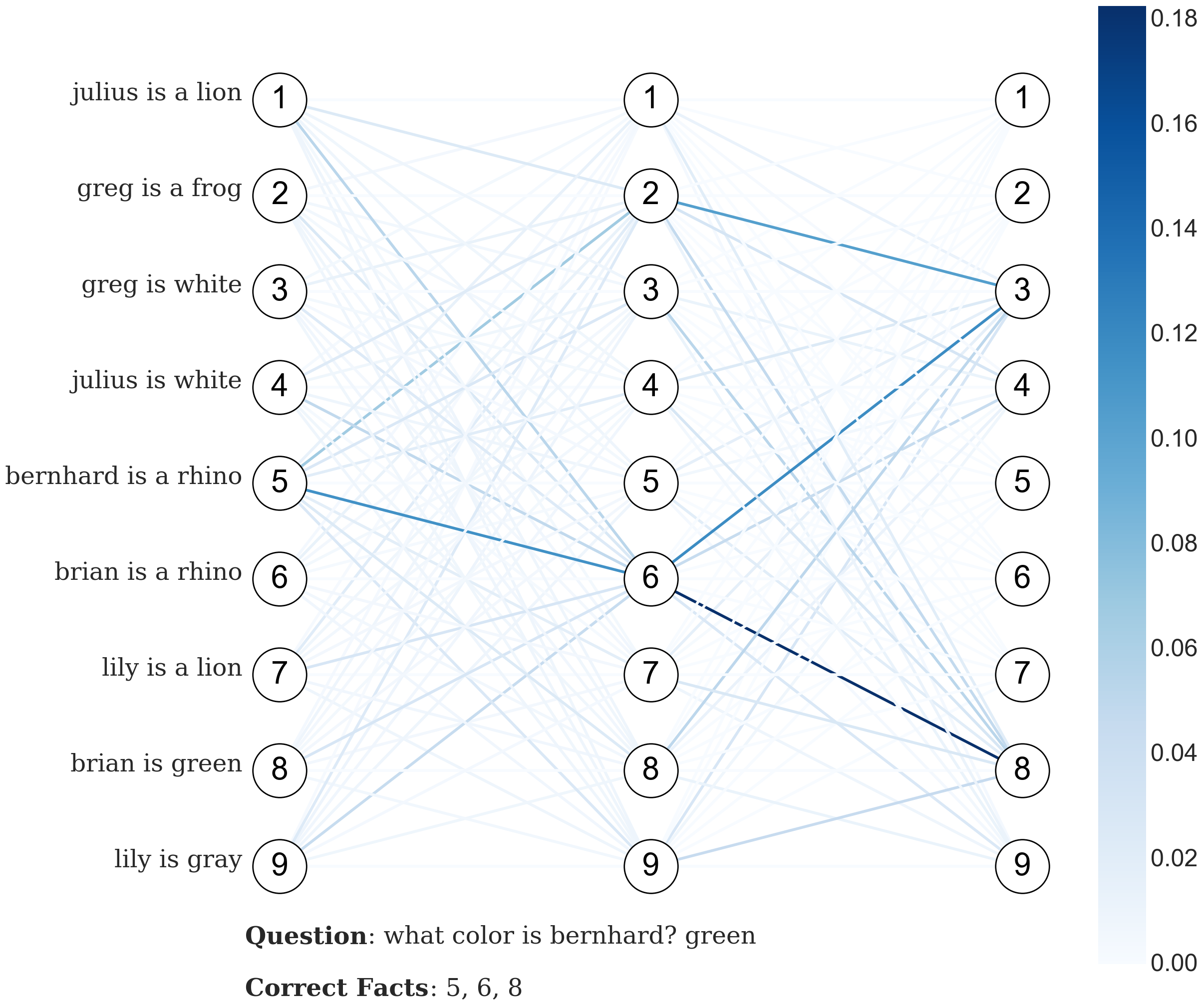}
\caption{\label{fig:vis4} \small Visualization of the attention distribution over supporting fact sequences for an example question in task $16$ for the Binary CRF model.
The actual question is displayed at the bottom along with the correct answer and the ground truth supporting facts ($5 \rightarrow 6 \rightarrow 8$).
The edges represent the marginal probabilities $p(z_k, z_{k+1} \given x, q)$, and the nodes represent the $n$ supporting facts (here we have $n = 9$).
The text for the supporting facts are shown on the left. The top three most likely sequences are:
$ p(z_1 = 5, z_2 = 6, z_3 = 8 \given x, q) = 0.0564, p(z_1 = 5, z_2 = 6, z_3 = 3 \given x, q) = 0.0364, p(z_1 = 5, z_2 = 2, z_3 = 3 \given x, q) = 0.0356$.}
\end{figure}

Our third experiment is on question answering (QA) with the linear-chain CRF attention layer for
inference over multiple facts.
We use the bAbI dataset \citep{Weston2015}, where the input is a set of sentences/facts paired
with a question, and the answer is a single token. For many of the tasks the model has 
to attend to multiple supporting facts to arrive at the correct answer (see Figure~\ref{fig:vis4} for an example),
and existing approaches use multiple `hops' to greedily attend to different facts. 
We experiment with employing structured attention to perform inference in a non-greedy way.
As the ground truth supporting facts are given in the dataset, we are able to assess the model's inference accuracy.

The baseline (simple) attention model is the End-To-End Memory Network \citep{Sukhbaatar2015} (MemN2N),
which we briefly describe here. See Appendix~\ref{app:qa} for full model details.
Let $\xvec_1, \dots, \xvec_n$ be the input embedding vectors for the $n$ sentences/facts and let
$\mathbf{q}$ be the
query embedding. In MemN2N, $z_k$ is the random variable for the sentence to
select at the $k$-th inference step (i.e. $k$-th hop), and thus $z_k \in \{1, \dots, n\}$.
The probability distribution over $z_k$ is given by
$p(z_k = i \given x, q) = \softmax((\xvec_i^k)^\top\qvec^k)$, and the context vector is given by
$\cvec^k = \sum_{i=1}^n p(z_k = i \given x, q) \mathbf{o}_i^k$, where $\xvec_i^k, \mathbf{o}_i^k$ are the 
input and output embedding for the $i$-th sentence at the $k$-th hop, respectively.
The $k$-th context vector is used to modify the query $\qvec^{k+1} = \qvec^k + \cvec^k$, and this 
process repeats for $k = 1, \dots, K$ (for $k=1$ we have $\xvec_i^k = \xvec_i, \qvec^k = \qvec, \cvec^k = \mathbf{0}$).
The $K$-th context and query vectors are used to obtain the final answer.
The attention mechanism for a 
$K$-hop MemN2N network can therefore be interpreted as a greedy selection of a length-$K$ 
sequence of facts (i.e. $z_1, \dots, z_K$).

For structured attention, we use an $n$-state, $K$-step linear-chain CRF.\footnote{Note that this 
differs from the segmentation attention for the neural machine translation experiments described above, which was a $K$-state (with $K =2$), $n$-step linear-chain CRF.}
 We experiment with two
different settings: (a) a unary CRF model with node potentials 
$$\theta_k(i) = (\xvec_i^k)^\top \mathbf{q}^k$$ and (b)
a binary CRF model with pairwise potentials 
$$\theta_{k,k+1}(i, j) = 
 (\mathbf{x}_i^k)^\top\qvec^k +  (\mathbf{x}_i^k)^\top \xvec_j^{k + 1}  +  (\mathbf{x}_j^{k + 1})^\top \mathbf{q}^{k + 1}$$
The binary CRF model is designed to test the model's ability to perform sequential reasoning. 
For both (a) and (b), a \emph{single} context vector is computed:
$\mathbf{c} = \sum_{z_1,\ldots,z_K} p(z_1,\ldots,z_K \given x,q) f(x,z)$ (unlike 
MemN2N which computes $K$ context vectors). Evaluating $\mathbf{c}$ 
requires summing over all $n^K$ possible sequences of length $K$, which may not be
practical for large values of $K$. However, if $f(x,z)$ factors over the components of $z$
(e.g. $f(x,z)= \sum_{k=1}^K f_k(x,z_k)$) then one can rewrite the above sum in terms of marginals:
$\mathbf{c} = \sum_{k=1}^K \sum_{i=1}^n p(z_{k} = i \given x,q) f_{k}(x,z_{k})$. In our experiments, we use
$f_k(x,z_k) = \mathbf{o}_{z_k}^k$. All three models are
described in further detail in Appendix~\ref{app:qa}. 

\begin{table}[t]
\small
\centering
\begin{tabular}{lcrrrrrr}
\toprule
&  & \multicolumn{2}{c}{MemN2N} & \multicolumn{2}{c}{Binary CRF} & \multicolumn{2}{c}{Unary CRF} \\
Task & $K$ & Ans $\%$& Fact $\%$  & Ans $\%$& Fact $\%$ & Ans $\%$& Fact $\%$ \\
 \midrule
 \textsc{Task 02 - Two Supporting Facts} &2& $87.3$ & $46.8$ & $84.7$ & $81.8$ & $43.5$ & $22.3$ \\
 \textsc{Task 03 - Three Supporting Facts} &3& $52.6$ & $1.4$ & $40.5$ & $0.1$ &$28.2$ & $0.0$\\
 \textsc{Task 07 - Counting} &3& $83.2$ &$-$ & $83.5$ & $-$ & $79.3$ & $-$\\
 \textsc{Task 08 - Lists Sets} &3& $94.1$ &$-$ & $93.3$ & $-$ & $87.1$ & $-$\\
 \textsc{Task 11 - Indefinite Knowledge} &2& $97.8$ & $38.2$ &$97.7$ &$80.8$  &$88.6$ & $0.0$ \\
 \textsc{Task 13 - Compound Coreference} &2& $95.6$ & $14.8$ &$97.0$ & $36.4$ &$94.4$ & $9.3$ \\
 \textsc{Task 14 - Time Reasoning} &2& $99.9$ & $77.6$ &$99.7$ & $98.2$  &$90.5$ & $30.2$\\
 \textsc{Task 15 - Basic Deduction} &2& $100.0$ & $59.3$  & $100.0$ & $89.5$ &$100.0$ & $51.4$\\
 \textsc{Task 16 - Basic Induction} &3& $97.1$ & $91.0$ &$97.9$ & $85.6$  &$98.0$ & $41.4$\\
 \textsc{Task 17 - Positional Reasoning} &2& $61.1$ & $23.9$ &$60.6$ & $49.6$ &$59.7$ & $10.5$\\
 \textsc{Task 18 - Size Reasoning} &2& $86.4$ & $3.3$ &$92.2$ & $3.9$ &$92.0$ & $1.4$\\
 \textsc{Task 19 - Path Finding} &2& $21.3$ & $10.2$ &$24.4$ & $11.5$ &$24.3$ & $7.8$ \\
\midrule
 \textsc{Average} &$-$& $81.4$ & $39.6$ & $ 81.0 $ & $53.7$ & $73.8$ & $17.4$\\
 \bottomrule
\end{tabular}
\caption{\label{tab:results} Answer accuracy (Ans $\%$) and supporting fact selection accuracy (Fact $\%$)
of the three QA models on the $1$K bAbI dataset. $K$ indicates the number of hops/inference steps used for each task.
Task 7 and 8 both contain variable number of facts and hence they are excluded from the fact accuracy
 measurement. Supporting fact selection accuracy is calculated by taking the average of $10$ best
runs (out of $20$) for each task.}
\end{table}

\paragraph{Results}
We use the version of the dataset with $1$K questions for each task. 
Since all models reduce to the same network for tasks with $1$ supporting fact, they are excluded from our experiments.
The number of hops (i.e. $K$) is task-dependent, and the number of memories (i.e. $n$) is limited to be at most $25$ (note
that many question have less than $25$ facts---e.g. the example in Figure~\ref{fig:vis4} has $9$ facts).
Due to high variance in model performance, we train $20$ models with different initializations for each task and report the 
test accuracy of the model that performed the best on a $10\%$ held-out validation set 
(as is typically done for bAbI tasks).

Results of the three different models are shown in Table~\ref{tab:results}. 
For correct answer seletion (Ans $\%$), we find that MemN2N and the Binary CRF model 
perform similarly while the Unary CRF model does worse, indicating the importance
of including pairwise potentials. We also assess each model's ability to attend to the correct supporting facts in 
Table~\ref{tab:results} (Fact $\%$).
Since ground truth supporting facts are provided for each query, we can check the sequence accuracy of supporting facts 
for each model  
(i.e. the rate of selecting the exact correct sequence of facts) by taking the highest probability sequence 
 $\hat{z} = \argmax p(z_1, \dots, z_K \given x, q)$ from the model and checking against the ground truth.
Overall the Binary CRF is able to recover supporting facts better than MemN2N. This improvement 
is significant and can be up to two-fold as seen for task $2$, $11$, $13$ \& $17$. 
However we observed that on many tasks it is sufficient to select only the last (or first) fact correctly to
predict the answer,
and thus higher sequence selection accuracy does not necessarily imply better answer accuracy (and vice versa).
For example, all three models get $100 \%$ answer accuracy on task $15$ but have different supporting fact accuracies.

Finally, in Figure~\ref{fig:vis4} we visualize of the output 
edge marginals produced by the Binary CRF model for a single question in task $16$. In this instance, the model
is uncertain but ultimately able to select the right sequence of facts $5 \rightarrow 6 \rightarrow 8$.

\subsection{Natural Language Inference}
The final experiment looks at the task of natural language inference
(NLI) with the syntactic attention layer. In NLI, the model is given two sentences (hypothesis/premise)
and has to predict their relationship: entailment, contradiction,
neutral.

For this task, we use the Stanford NLI dataset \citep{Bowman2015} and model
our approach off of the decomposable attention model of
\cite{Parikh2016}. This model
 takes in the matrix of word embeddings as the input for each sentence and performs \textit{inter-sentence}
attention to predict the answer.
Appendix~\ref{app:nli} describes the full model.

As in the transduction task, we focus on modifying the input representation 
to take into account soft parents via self-attention (i.e. \textit{intra-sentence} attention).
In addition to the three baselines
described for tree transduction (No Attention, Simple, Structured), we
also explore two additional settings: (d) \textit{hard} pipeline parent selection,  
i.e. $\hat{\mathbf{x}}_j = [\mathbf{x}_j; \mathbf{x}_{\head(j)}]$, 
where $\head(j)$ is the index of $x_j$'s parent\footnote{The parents are obtained from running the
 dependency parser of \cite{Andor2016}, available at \\ \url{https://github.com/tensorflow/models/tree/master/syntaxnet}}; 
(e) \textit{pretrained} structured attention:  structured attention where the parsing layer is
pretrained for one epoch on a parsed dataset (which was enough for convergence).

\begin{table}
\small
\center
\begin{tabular}{l c}
\toprule
Model &  Accuracy $\%$ \\
\midrule
Handcrafted features \citep{Bowman2015} &  $78.2$ \\
LSTM encoders \citep{Bowman2015} &  $80.6$ \\
Tree-Based CNN \citep{Mou2016} & $82.1$ \\
Stack-Augmented Parser-Interpreter Neural Net \citep{Bowman2016} &  $83.2$ \\
LSTM with word-by-word attention \citep{Rock2016} &  $83.5$ \\
Matching LSTMs \citep{Wang2016} &  $86.1$ \\
Decomposable attention over word embeddings \citep{Parikh2016}  & $86.3$ \\
Decomposable attention $+$ intra-sentence attention \citep{Parikh2016} & $86.8$ \\
Attention over constituency tree nodes \citep{Zhao2016}  & $87.2$ \\
Neural Tree Indexers \citep{Munkhdalai2016}  & $87.3$ \\
Enhanced BiLSTM Inference Model \citep{Chen2016}  & $87.7$ \\
Enhanced BiLSTM Inference Model $+$ ensemble \citep{Chen2016} & $88.3$ \\
\midrule
No Attention & $85.8$  \\
No Attention + Hard parent & $86.1$  \\
Simple Attention & $86.2 $ \\
Structured Attention & $86.8$ \\
Pretrained Structured Attention  & $86.5$  \\
\bottomrule
\end{tabular}
\caption{\label{tab:main} \small Results of our models (bottom) and others (top) on the Stanford NLI test set.
Our baseline model has the same architecture as \cite{Parikh2016} but the performance is
slightly different due to different settings (e.g. we train for $100$ epochs
with a batch size of $32$ while \cite{Parikh2016} train for
$400$ epochs with a batch size of $4$ using asynchronous SGD.)}
\end{table}

\paragraph{Results}
Results of our models are shown in Table~\ref{tab:main}. Simple attention improves upon the no attention model, and this is
consistent with improvements observed by \cite{Parikh2016} with their intra-sentence attention model. The pipelined model with
hard parents also slightly improves upon the baseline. Structured attention outperforms both models, though surprisingly, pretraining the syntactic attention
layer on the parse trees performs worse than training it from scratch---it is possible that the
pretrained attention is too strict for this task. 

We also obtain the hard parse for an example sentence by running the
Viterbi algorithm on the syntactic attention layer with the non-pretrained model:
\begin{center}
\includegraphics[scale=0.8]{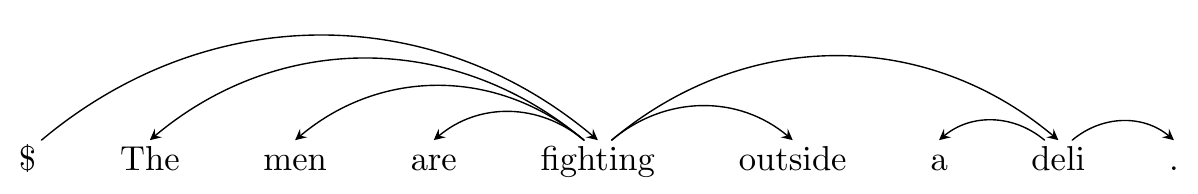}
\end{center}
Despite being trained without ever being exposed to an explicit parse tree, the syntactic attention
layer learns an almost plausible dependency structure. In the above example it is able to correctly
identify the main verb \texttt{fighting}, but makes mistakes  on determiners
(e.g. head of \texttt{The} should be \texttt{men}). We generally observed this pattern across sentences, 
possibly because the verb structure is more important for the inference task. 

\section{Conclusion}

This work outlines structured attention networks, which incorporate
graphical models to generalize simple attention, and describes the
technical machinery and computational techniques for backpropagating
through models of this form. We implement two classes of structured
attention layers: a linear-chain CRF (for neural machine translation 
and question answering) and a more complicated first-order dependency
parser (for tree transduction and natural language inference). Experiments show that this method can learn interesting
structural properties and improve on top of standard models. 
Structured attention could also be a way of learning latent labelers or parsers through
attention on other tasks.

It should be noted that the 
additional complexity in computing the attention distribution
increases run-time---for example, 
structured attention was approximately $5 \times$ slower
to train than simple attention for the neural machine translation
experiments, even though both attention layers have the same asymptotic run-time (i.e. $O(n)$). 

Embedding \textit{differentiable inference} (and more generally, \textit{differentiable algorithms}) 
into deep models is an exciting
area of research. While we have focused on models that admit 
(tractable) exact inference, similar technique can be used to embed
approximate inference methods. 
Many optimization algorithms (e.g. gradient descent, LBFGS) are also differentiable \citep{domke2012generic,Maclaurin2015},
and have been used as output layers for structured prediction in energy-based models
 \citep{Belanger2016,wang2016nips}.
Incorporating them as internal neural network layers is an interesting avenue for future work.

\subsubsection*{Acknowledgments}

We thank Tao Lei, Ankur Parikh, Tim Vieira, Matt Gormley, Andr{\'e} Martins, Jason Eisner, Yoav Goldberg, and the anonymous reviewers 
for helpful comments, discussion, notes, and code. We additionally thank Yasumasa Miyamoto for verifying Japanese-English translations.

\bibliography{master.bib}
\bibliographystyle{iclr2017_conference}

\newpage
\appendix
\section*{APPENDICES}
\section{Model Details}\label{app:model}
\subsection{Syntactic Attention}\label{app:parsing}
The syntactic attention layer (for tree transduction and natural language inference) is similar to the first-order graph-based 
dependency parser of \cite{Kipperwasser2016}.
Given an input sentence $[x_1, \dots, x_n]$ and the corresponding word vectors
$[\xvec_1, \dots, \xvec_n]$, we use a bidirectional LSTM
to get the hidden states for each time step $i \in [1, \dots, n]$,
\begin{align*}
\hvec_i^{\text{fwd}} = \lstm(\xvec_i, \hvec_{i-1}^{\text{fwd}}) & & \hvec_i^{\text{bwd}} = \lstm(\xvec_i, \hvec_{i+1}^{\text{bwd}}) & & \hvec_i =  [\hvec_i^{\text{fwd}} ; \hvec_i^{\text{bwd}}] 
\end{align*}
where the forward and backward LSTMs have their own parameters. The score for 
$x_i \rightarrow x_j$ (i.e. $x_i$ is the parent of $x_j$), is given by an MLP
\begin{equation*}
\theta_{ij} = \tanh( \svec^\top\tanh(\Wvec_1 \hvec_i + \Wvec_2 \hvec_j + \bvec))
\end{equation*}
These scores are used as input to the inside-outside algorithm (see Appendix~\ref{app:io})
 to obtain the probability
of each word's parent $p(z_{ij} = 1 \given x)$, which is used to obtain the 
soft-parent $\cvec_j$ for each word $x_j$. In the non-structured case 
we simply have $p(z_{ij} = 1 \given x) = \softmax(\theta_{ij})$.

\subsection{Tree Transduction}\label{app:tree}
Let $[x_1, \dots, x_n],[y_1, \dots, y_m]$ be the sequence of source/target
 symbols, with the associated embeddings $[\xvec_1, \dots, \xvec_n],
[\yvec_1, \dots, \yvec_m]$ with $\xvec_i, \yvec_j \in \reals^l$. In the simplest baseline
model we take the source representation to be the matrix of the
symbol embeddings. The decoder is a one-layer LSTM which produces the hidden states
$\hvec_j' = \lstm(\yvec_j, \hvec_{j-1}')$, with $\hvec_j' \in \reals^l$.
The hidden states are combined with the input representation via a bilinear map
$\Wvec \in \reals^{l \times l}$ to produce the attention distribution 
used to obtain the vector $\mvec_i$, which is combined with the decoder hidden state
as follows,
\begin{align*}
\alpha_i = \frac{\exp \xvec_i \Wvec \hvec_j'}{\sum_{k=1}^n \exp \xvec_k \Wvec \hvec_j'}
& & \mvec_i = \sum_{i=1}^n \alpha_i \xvec_i & & \hat{\hvec}_j = \tanh (\Uvec [\mvec_i ; \hvec_j'] )
\end{align*}
Here we have $\Wvec \in \reals^{l \times l}$ and $\Uvec \in \reals^{2l \times l}$. Finally, $\hat{\hvec}_j$ is used to to obtain a distribution
over the next symbol $y_{j+1}$,
\begin{equation*}
p(y_{j+1} \given x_1, \dots, x_n, y_1, \dots, y_j) = \softmax(\Vvec \hat{\hvec}_j + \bvec)
\end{equation*}
For structured/simple models, the $j$-th source representation are respectively
\begin{align*}
\hat{\xvec}_i = \left[\xvec_i ; \sum_{k=1}^n p(z_{ki} = 1 \given x ) \, \xvec_k\right] & &\hat{\xvec}_i = \left[\xvec_i ; \sum_{k=1}^n \softmax (\theta_{ki})\, \xvec_k\right]
\end{align*}
where $\theta_{ij}$ comes from the bidirectional LSTM described in~\ref{app:parsing}.
Then $\alpha_i$ and $\mvec_i$ changed accordingly,
\begin{align*}
\alpha_i = \frac{\exp \hat{\xvec}_i \Wvec \hvec_j'}{\sum_{k=1}^n \exp \hat{\xvec}_k \Wvec \hvec_j'}  & &
\mvec_i = \sum_{i=1}^n \alpha_i \hat{\xvec}_i
\end{align*}
Note that in this case we have $\Wvec \in \reals^{2l \times l}$ and $\Uvec \in \reals^{3l \times l}$.
We use $l = 50$ in all our experiments. The forward/backward LSTMs for the parsing LSTM 
are also $50$-dimensional. Symbol embeddings are shared between the encoder and the parsing LSTMs.

Additional training details include: batch size of $20$; training for 
$13$ epochs with a learning rate of $1.0$, which starts decaying by half after epoch $9$
(or the epoch at which performance does not improve on validation, whichever comes first);
parameter initialization over a uniform distribution $U[-0.1, 0.1]$; gradient normalization at $1$
(i.e. renormalize the gradients to have norm $1$ if the $l_2$ norm exceeds $1$). Decoding
is done with beam search (beam size $ = 5$).

\subsection{Neural Machine Translation}\label{app:nmt}
The baseline NMT system is from \cite{Luong2015}.
Let $[x_1, \dots, x_n],[y_1, \dots, y_m]$ be the 
source/target sentence, with the associated word embeddings $[\xvec_1, \dots, \xvec_n],
[\yvec_1, \dots, \yvec_m]$. 
The encoder is an LSTM over the source sentence, which produces the hidden states $[\hvec_1, \dots, \hvec_n]$
where 
\begin{equation*}
\hvec_i = \lstm(\xvec_i, \hvec_{i-1})
\end{equation*}
 and $\hvec_i \in \reals^l$. 
The decoder is another LSTM 
which produces the hidden states $\hvec_j' \in \reals^l$.
In the simple attention case with categorical attention, the hidden states 
are combined with the input representation via a bilinear map
$\Wvec \in \reals^{l \times l}$ and this distribution is used to obtain the context vector
at the $j$-th time step,
\begin{align*}
\theta_i = \hvec_i \Wvec \hvec_j' & & \cvec_j = \sum_{i=1}^n \softmax(\theta_i)\hvec_i
\end{align*} 
The Bernoulli attention network has the same $\theta_i$ but instead uses a $\sigmoid$ to obtain the weights of the linear combination, i.e.,
\begin{align*}
\cvec_j = \sum_{i=1}^n \sigmoid(\theta_i) \hvec_i
\end{align*}
And finally, the structured attention model uses a bilinear map to parameterize one of the unary potentials
\[
    \theta_i(k)= 
\begin{cases}
    \hvec_i \Wvec \hvec_j',& k = 1           \\ 0,  &k = 0
\end{cases}
\]
\begin{align*}
 \theta_{i,i+1}(z_i, z_{i+1}) &= \theta_i(z_i) + \theta_{i+1}(z_{i+1})
+ \mathbf{b}_{z_i, z_{i+1}}
\end{align*}
where $\bvec$ are the pairwise potentials. 
These potentials are used as inputs to the forward-backward algorithm to obtain the marginals $p(z_i = 1 \given x, q)$,
which are further normalized to obtain the context vector
\begin{align*}
\cvec_j = \sum_{i=1}^n \frac{p(z_i=1 \given x, q)}{\gamma} \hvec_i  & & \gamma = \frac{1}{\lambda} \sum_i^n p(z_i =1 \given x, q)  
\end{align*}
We use $\lambda = 2$ and also add an $l_2$ penalty of $0.005$ on the pairwise potentials $\bvec$.
The context vector is then combined with the decoder hidden state
\begin{align*}
\hat{\hvec}_j = \tanh (\Uvec[\cvec_j ; \hvec_j']) 
\end{align*}
and $\hat{\hvec}_j$ is used to obtain the distribution over the next target word $y_{j+1}$
\begin{align*}
p(y_{j+1} \given x_1, \dots, x_n, y_1, \dots y_j) = \softmax(\Vvec \hat{\hvec}_j + \bvec)
\end{align*}

The encoder/decoder LSTMs have $2$ layers and $500$  hidden units (i.e. $l = 500$).

Additional training details include: batch size of $128$; training for 
$30$ epochs with a learning rate of $1.0$, which starts decaying by half 
after the first epoch at which performance does not improve on validation;
dropout with probability $0.3$;
parameter initialization over a uniform distribution $U[-0.1, 0.1]$; gradient normalization at $1$.
We generate target translations with beam search (beam size $= 5$), and evaluate with \texttt{multi-bleu.perl} from Moses.\footnote{
\url{https://github.com/moses-smt/mosesdecoder/blob/master/scripts/generic/multi-bleu.perl}}
\subsection{Question Answering}\label{app:qa}
Our baseline model (MemN2N) is implemented following the same architecture as described in \cite{Sukhbaatar2015}. In particular,
let $x = [x_1, \dots, x_n]$ represent the sequence of $n$ facts with the associated embeddings $[\mathbf{x}_1, \dots, \xvec_n]$ 
and let $\qvec$ be the embedding of the query $q$. The
embeddings are obtained by simply adding the word embeddings in each sentence or query.
The full model with $K$ hops is as follows:
\begin{align*}
  &p(z_k = i \given x, q) = \softmax( (\mathbf{x}_i^k)^\top \mathbf{q}^k ) \\
  &\mathbf{c}^k = \sum_{i=1}^n p(z_k = i \given x, q) \mathbf{o}_i^k \\
  &\mathbf{q}^{k + 1} = \mathbf{q}^k + \mathbf{c}^k \\
  &p(y \given x, q) = \softmax(\Wvec (\mathbf{q}^K + \mathbf{c}^K))
\end{align*}
where $p(y \given x, q)$ is the distribution over the answer vocabulary. At each layer, $\{\mathbf{x}_i^k\}$ and $\{\mathbf{o}_i^k\}$
are computed using embedding matrices $\mathbf{X}^k$ and $\mathbf{O}^k$. We use the \emph{adjacent weight tying scheme} from the paper
so that $\mathbf{X}^{k+1} = \mathbf{O}^k, \mathbf{W}^T = \mathbf{O}^K$. $\mathbf{X}^1$ is also used to compute the query embedding at the first hop. 
For $k=1$ we have $\xvec_i^k = \xvec_i, \qvec^k = \qvec, \cvec^k = \mathbf{0}$.

For both the Unary and the Binary CRF models, the same input fact and query representations are computed (i.e. same embedding matrices
with weight tying scheme). For the unary model, the potentials are parameterized as
\[ \theta_{k}(i) = (\xvec_i^k)^\top \mathbf{q}^k \]

and for the binary model we compute pairwise potentials as
\[ \theta_{k,k+1}(i, j) =  (\mathbf{x}_i^k)^\top\qvec^k +  (\mathbf{x}_i^k)^\top \xvec_j^{k + 1}  +  (\mathbf{x}_j^{k + 1})^\top \mathbf{q}^{k + 1} \]
The $\qvec^k$'s are updated simply with a linear mapping, i.e.
\[ \mathbf{q}^{k+1} = \mathbf{Q} \mathbf{q}^k \]
In the case of the Binary CRF, to discourage the model from selecting the same fact again we additionally set $\theta_{k,k+1}(i,i) = -\infty$ for all $i \in \{1, \dots, n\}$. 
Given these potentials, we compute the marginals $p(z_k = i, z_{k+1} = j \given x, q)$ using the forward-backward algorithm, which is then used to compute the context vector:
\begin{align*}
\mathbf{c} = \sum_{z_1,\ldots,z_K} p(z_1,\ldots,z_K \given x,q) f(x,z) & &  f(x,z) = \sum_{k=1}^K f_k(x, z_k) & & f_k(x,z_k) = \mathbf{o}_{z_k}^k 
\end{align*}
Note that if $f(x,z)$ factors over the components of $z$ (as is the case above) then computing $\cvec$ only requires evaluating the marginals
$p(z_k \given x,q)$.

Finally, given the context vector the prediction is made in a similar fashion to MemN2N:
\begin{align*}
p(y \given x, q) = \softmax(\Wvec (\mathbf{q}^K + \mathbf{c}))
\end{align*}

Other training setup is similar to \cite{Sukhbaatar2015}: we use stochastic gradient descent with 
learning rate $0.01$, which is divided by $2$ every $25$ epochs until $100$ epochs are reached. Capacity of 
the memory is limited to $25$ sentences. The embedding vectors are of size $20$ and gradients are renormalized
if the norm exceeds $40$. All models implement \emph{position encoding}, \emph{temporal encoding}, and \emph{linear start} from
the original paper. For linear start, the $\softmax(\cdot)$ function in the attention layer
is removed at the beginning and re-inserted after $20$ epochs for MemN2N, while for the CRF models we apply a $\log(\softmax(\cdot))$ 
layer on the $\qvec^k$ after $20$ epochs. Each model is trained separately for each task. 

\subsection{Natural Language Inference}\label{app:nli}
Our baseline model/setup is essentially the same as that of \cite{Parikh2016}. Let $[x_1, \dots, x_n], [y_1, \dots, y_m]$
be the premise/hypothesis, with the corresponding 
input representations $[\xvec_1, \dots, \xvec_n], [\yvec_1, \dots, \yvec_m]$.
The input representations are obtained by a linear transformation of the $300$-dimensional pretrained GloVe embeddings \citep{Pennington2014}
after normalizing the GloVe embeddings to have unit norm.\footnote{We use
the GloVe embeddings pretrained over the $840$ billion word Common Crawl, publicly available at \url{http://nlp.stanford.edu/projects/glove/}} 
The pretrained embeddings remain fixed but the linear layer (which is also $300$-dimensional) is trained. Words not in the pretrained vocabulary
are hashed to one of $100$ Gaussian embeddings with mean $0$ and standard deviation $1$.

We concatenate each input representation with a convex combination of the other sentence's input representations 
(essentially performing \textit{inter-sentence} attention),
where the weights are determined through a dot product followed by a softmax,  
\begin{align*}
e_{ij} = f(\xvec_i)^\top f(\yvec_j) & & \bar{\xvec}_{i} = \left[\xvec_i ; \sum_{j=1}^m \frac{\exp e_{ij}}{\sum_{k=1}^m \exp e_{ik}} \yvec_{j}\right] & & 
\bar{\yvec}_{j} = \left[\yvec_j ; \sum_{i=1}^n \frac{\exp e_{ij}}{\sum_{k=1}^n \exp e_{kj}} \xvec_{i}\right]
\end{align*}
Here $f(\cdot)$ is an MLP. The new representations are fed through another MLP $g(\cdot)$, summed,
combined with the final MLP $h(\cdot)$ and fed through a softmax layer to obtain a distribution
over the labels $l$,
\begin{align*}
\bar{\xvec} &= \sum_{i=1}^n g(\bar{\xvec}_{i}) \hspace{20mm} \bar{\yvec} = \sum_{j=1}^m g(\bar{\yvec}_{j}) \\
p(l \given x_1, &\dots, x_n, y_1, \dots, y_m)= \softmax(\Vvec h([\bar{\xvec}; \bar{\yvec}]) + \bvec)
\end{align*}

All the MLPs have $2$-layers, $300$ $\relu$ units, and dropout probability of $0.2$. For structured/simple models,
we first employ the bidirectional parsing LSTM (see \ref{app:parsing}) to obtain the scores $\theta_{ij}$. In the structured case
each word representation is simply concatenated with its soft-parent
\begin{align*}
\hat{\xvec}_i = \left[\xvec_i ; \sum_{k=1}^n p(z_{ki} = 1 \given x ) \xvec_k\right] % & & \hat{\yvec}_j = [\yvec_j ; \sum_{l=1}^m p(z'_{lj} = 1 \given y) \yvec_l]
\end{align*}
and $\hat{\xvec}_i$ (and analogously $\hat{\yvec}_j$) is used as the input to the above model. In the simple case (which
closely corresponds to the \emph{intra-sentence} attention model of \cite{Parikh2016}), we have
\begin{align*}
\hat{\xvec}_i = \left[\xvec_i ; \sum_{k=1}^n \frac{\exp \theta_{ki}}{\sum_{l=1}^n \exp \theta_{li}} \xvec_k \right]
\end{align*}
The word embeddings for the parsing LSTMs are also initialized with GloVe, and the parsing layer is shared between the two sentences.
The forward/backward LSTMs for the parsing layer are $100$-dimensional.

Additional training details include: batch size of $32$; training for $100$ epochs with Adagrad \citep{Duchi2011} where the global
learning rate is $0.05$ and sum of gradient squared is initialized to $0.1$;
parameter intialization over a Gaussian distribution with mean $0$ and standard deviation $0.01$;
gradient normalization at $5$. In the pretrained scenario, pretraining is done with Adam \citep{Kingma2015} with
learning rate equal to $0.01$, and $\beta_1 = 0.9$, $\beta_2 = 0.999$.

\section{Forward/Backward through the Inside-Outside Algorithm}\label{app:io}
Figure~\ref{fig:io-fprop} shows the procedure for obtaining the parsing marginals from
the input potentials. This corresponds to running the inside-outside version of Eisner's algorithm \citep{Eisner1996}.
The intermediate data structures used during the dynamic programming algorithm
are the (log) inside tables $\alpha$, and the (log) outside tables $\beta$. Both $\alpha, \beta$
are of size $n \times n \times 2 \times 2$, where $n$ is the sentence length.
 First two dimensions encode the start/end index
of the span (i.e. subtree). The third dimension encodes whether the root of the subtree is
the left ($L$) or right ($R$) index of the span. The fourth dimension indicates if the span is complete ($1$)
or incomplete ($0$). We can calculate the marginal distribution of each word's parent (for all words) in $O(n^3)$ using
this algorithm.

Backward pass through the inside-outside algorithm is slightly more involved, but still takes $O(n^3)$ time.
Figure~\ref{fig:io-bprop} illustrates the backward procedure, which receives the gradient of the loss
$\mcL$ with respect to the marginals, $\nabla^\mcL_p$, and computes the gradient of the loss with respect
to the potentials $\nabla^\mcL_\theta$. The computations must be performed in the signed log-space semifield
to handle log of negative values. See section~\ref{sec:e2e} and 
    Table~\ref{tab:dlog} for more details.

\begin{figure}
\small
    \begin{algorithmic}  
      \Procedure{InsideOutside}{$\theta$} 
\State{$\alpha, \beta \gets -\infty$} \Comment{Initialize log of inside ($\alpha$), outside ($\beta$) tables}
\For{$i = 1,\dots, n$}
\State{$\alpha[i, i, L, 1] \gets 0$}
\State{$\alpha[i, i, R, 1] \gets 0$}
\EndFor
\State{$\beta[1, n, R, 1] \gets 0$}
\For{$k = 1, \dots, n$} \Comment{Inside step}
\For{$s = 1, \dots, n-k$} 
\State{$t \gets s+k$}
\State{$\alpha[s,t, R, 0] \gets \bigoplus_{u \in [s, t-1]} \alpha[s, u, R, 1] \otimes \alpha[u+1,t,L,1] \otimes \theta_{st}$ }
\State{$\alpha[s,t, L, 0] \gets \bigoplus_{u \in [s, t-1]} \alpha[s, u, R, 1] \otimes \alpha[u+1,t,L,1] \otimes \theta_{ts}$ }
\State{$\alpha[s,t, R, 1] \gets \bigoplus_{u \in [s+1, t]} \alpha[s, u, R, 0] \otimes \alpha[u,t,R,1]$ }
\State{$\alpha[s,t, L, 1] \gets \bigoplus_{u \in [s, t-1]} \alpha[s, u, L, 1] \otimes \alpha[u,t,L,0]$ }
\EndFor
\EndFor
\For{$k = n, \dots, 1 $} \Comment{Outside step}
\For{$s = 1, \dots, n-k$} 
\State{$t \gets s+k$}
\For{$u = s+1, \dots, t$}
\State{$\beta[s,u , R, 0] \oplusgets \beta[s, t, R, 1] \otimes \alpha[u,t,R,1]$ }
\State{$\beta[u,t, R, 1] \oplusgets \beta[s, t, R, 1] \otimes \alpha[s,u,R,0]$ }
\EndFor
\If{$s > 1$}
\For{$u = s, \dots, t-1$}
\State{$\beta[s,u , L, 1] \oplusgets \beta[s, t, L, 1] \otimes \alpha[u,t,L,0]$ }
\State{$\beta[u,t, L, 0] \oplusgets \beta[s, t, L, 1] \otimes \alpha[s,u,L,1]$ }
\EndFor
\EndIf
\For{$u = s, \dots, t-1$}
\State{$\beta[s,u , R, 1] \oplusgets \beta[s, t, R, 0] \otimes \alpha[u+1,t,L,1] \otimes \theta_{st}$ }
\State{$\beta[u+1,t, L, 1] \oplusgets \beta[s, t, R, 0] \otimes \alpha[s,u,R,1] \otimes \theta_{st}$ }
\EndFor
\If{$s > 1$}
\For{$u = s, \dots, t-1$}
\State{$\beta[s,u , R, 1] \oplusgets \beta[s, t, L, 0] \otimes \alpha[u+1,t,L,1] \otimes \theta_{ts}$ }
\State{$\beta[u+1,t, L, 1] \oplusgets \beta[s, t, L, 0] \otimes \alpha[s,u,R,1] \otimes \theta_{ts}$ }
\EndFor
\EndIf

\EndFor
\EndFor
\State{$A \gets \alpha[1, n, R, 1]$} \Comment{Log partition}
\For{$s = 1, \dots, n-1$}  \Comment{Compute marginals. Note that $p[s,t] = p(z_{st} = 1 \given x)$}
\For{$t = s+1, \dots, n$}
\State{$p[s,t] \gets \exp (\alpha[s,t,R, 0] \otimes \beta[s,t,R,0] \otimes - A )$} 
\If{$s > 1$}
\State{$p[t,s] \gets \exp (\alpha[s,t,L, 0] \otimes \beta[s,t,L,0] \otimes - A )$}
\EndIf
\EndFor
\EndFor
      \State{\Return{$p$}} 
      \EndProcedure{}
    \end{algorithmic}
  \caption{\label{fig:io-fprop} \small  Forward step of the syntatic attention layer to compute the marginals, using the inside-outside algorithm
\citep{Baker1979} on the data structures of \cite{Eisner1996}. 
We assume the special root symbol is the first element of the sequence, and that the sentence length is $n$.
Calculations are performed in log-space semifield with $\oplus = \logadd$  and $\otimes = +$ for numerical precision. 
$a,b \gets c$ means $a \gets c$ and $b \gets c$.
$a \oplusgets b$ means $a \gets a \oplus b$.}
\end{figure}

\begin{figure}
\small
    \begin{algorithmic}  
      \Procedure{BackpropInsideOutside}{$\theta, p, \pgrad$}
\For{$s,t = 1,\dots , n; s \ne t$} \Comment{Backpropagation uses the identity $\nabla^\mcL_{\theta} = (p \odot \pgrad) \nabla^{\log p}_{\theta}$}
\State{$\delta[s,t] \gets \log p[s,t] \otimes \log \pgrad[s,t]$} \Comment{$\delta = \log (p \odot \pgrad)$} 
\EndFor
\State{$\alphagrad, \betagrad, \thetagrad \gets -\infty$} \Comment{Initialize inside ($\alphagrad$), outside ($\betagrad$) gradients, and log of $\nabla^\mcL_\theta$}
\For{$s = 1, \dots, n-1$} \Comment{Backpropagate $\delta$ to  $\alphagrad$ and $\betagrad$}
\For{$t = s+1, \dots, n$}
\State{$\alphagrad[s,t,R, 0], \betagrad[s,t,R, 0] \gets \delta[s,t]$}
\State{$\alphagrad[1,n,R,1] \oplusgets    -\delta [s,t]$}
\If{$s > 1$}
\State{$\alphagrad[s,t,L, 0], \betagrad[s,t,L, 0] \gets \delta[t,s]$}
\State{$\alphagrad[1,n,R,1] \oplusgets   -\delta [s,t]$}
\EndIf
\EndFor
\EndFor
\For{$k = 1, \dots, n$} \Comment{Backpropagate through outside step}
\For{$s = 1, \dots, n-k$} 
\State{$t \gets s+k$}
\State{$\nu \gets \betagrad[s,t,R,0] \otimes \beta[s, t, R, 0]$} \Comment{$\nu, \gamma$ are temporary values}
\For{$u = t, \dots, n$}
\State{$\betagrad[s,u,R,1], \alphagrad[t,u,R,1] \oplusgets \nu \otimes \beta[s,u,R,1] \otimes \alpha[t, u, R, 1]$}
\EndFor
\If{$s > 1$}
\State{$\nu \gets \betagrad[s,t,L,0] \otimes \beta[s, t, L, 0]$}
\For{$u = 1, \dots, s$}
\State{$\betagrad[u,t,L,1],\alphagrad[u,s,L,1]  \oplusgets \nu \otimes \beta[u,t,L,1] \otimes \alpha[u, s, L, 1]$}
\EndFor
\State{$\nu \gets \betagrad[s,t,L,1] \otimes \beta[s,t,L,1]$}
\For{$u = t, \dots, n$}
\State{$\betagrad[s,u,L,1],\alphagrad[t,u,L,0] \oplusgets \nu \otimes \beta[s,u,L,1]\otimes \alpha[t,u,L,1]$}
\EndFor
\For{$u = 1, \dots, s-1$}
\State{$\gamma \gets \beta[u,t,R,0]\otimes \alpha[u,s-1,R,1]\otimes \theta_{ut}$}
\State{$\betagrad[u,t,R,0],\alphagrad[u,s-1,R,1],\thetagrad[u,t]  \oplusgets \nu \otimes \gamma$}
\State{$\gamma \gets \beta[u,t,L,0]\otimes \alpha[u,s-1,R,1]\otimes \theta_{tu}$}
\State{$\betagrad[u,t,L,0],\alphagrad[u,s-1,R,1],\thetagrad[t,u] \oplusgets \nu \otimes \gamma$}
\EndFor
\EndIf
\State{$\nu \gets \betagrad[s,t,R,1] \otimes \beta[s,t,R,1]$}
\For{$u =  1,\dots, s$}
\State{$\betagrad[u,t,R,1],\alphagrad[u,s,R,0]  \oplusgets \nu \otimes \beta[u,t,R,1]\otimes \alpha[u,s,R,0]$}
\EndFor
\For{$u = t+1, \dots, n$}
\State{$\gamma \gets \beta[s,u,R,0]  \otimes \alpha[t+1,u,L,1]\otimes \theta_{su}$}
\State{$\betagrad[s,u,R,0],\alphagrad[t+1,u,L,1],\thetagrad[s,u] \oplusgets \nu \otimes \gamma$}
\State{$\gamma \gets \beta[s,u,L,0]  \otimes \alpha[t+1,u,L,1]\otimes \theta_{us}$}
\State{$\betagrad[s,u,L,0],\alphagrad[t+1,u,L,1],\thetagrad[u,s] \oplusgets \nu \otimes \gamma$}
\EndFor
\EndFor
\EndFor
\For{$k = n, \dots, 1 $} \Comment{Backpropagate through inside step}
\For{$s = 1,\dots, n-k$} 
\State{$t \gets s+k$}
\State{$\nu \gets \alphagrad[s,t,R,1] \otimes \alpha[s,t,R,1]$}
\For{$u = s+1,\dots, t$}
\State{$\alphagrad[u,t,R,0],\alphagrad[u,t,R,1] \oplusgets \nu \otimes \alpha[s,u,R,0] \otimes \alpha[u,t,R,1]$}
\EndFor
\If{$s > 1$}
\State{$\nu \gets \alphagrad[s,t,L,1] \otimes \alpha[s,t,L,1]$}
\For{$u = s, \dots, t-1$}
\State{$\alphagrad[s,u,L,1],\alphagrad[u,t,L,0] \oplusgets \nu \otimes \alpha[s,u,L,1] \otimes \alpha[u,t,L,0]$}
\EndFor
\State{$\nu \gets \alphagrad[s,t,L,0] \otimes \alpha[s,t,L,0]$}
\For{$u = s,\dots, t-1$}
\State{$\gamma \gets \alpha[s,u,R,1] \otimes \alpha[u+1,t,L,1] \otimes \theta_{ts}$}
\State{$\alphagrad[s,u,R,1],\alphagrad[u+1,t,L,1],\thetagrad[t,s] \oplusgets \nu \otimes \gamma$}
\EndFor
\EndIf
\State{$\nu \gets \alphagrad[s,t,R,0] \otimes \alpha[s,t,R,0]$}
\For{$u = s,\dots, t-1$}
\State{$\gamma \gets \alpha[s,u,R,1] \otimes \alpha[u+1,t,L,1] \otimes \theta_{st}$ }
\State{$\alphagrad[s,u,R,1],\alphagrad[u+1,t,L,1],\thetagrad[s,t] \oplusgets \nu \otimes \gamma $}
\EndFor
\EndFor
\EndFor
\State{\Return{$\signexp \thetagrad$}}\Comment{Exponentiate log gradient, multiply by sign,  and return $\nabla^\mcL_\theta$}
\EndProcedure{}
    \end{algorithmic}
  \caption{\label{fig:io-bprop} \small Backpropagation through the inside-outside algorithm to calculate the gradient with respect
to the input potentials. $\nabla^a_b$ denotes
the Jacobian of $a$ with respect to $b$ (so $\nabla^\mcL_\theta$ is the gradient with respect to $\theta$).
 $a,b \oplusgets c$ means $a \gets a \oplus c$ and $b \gets b \oplus c$.}
\end{figure}

\end{document}